\documentclass{article}

\PassOptionsToPackage{sort&compress}{natbib} % Added by Sinan
\usepackage[main, final]{neurips_2026}
\usepackage[utf8]{inputenc} % allow utf-8 input
\usepackage[T1]{fontenc}    % use 8-bit T1 fonts
\usepackage{hyperref}       % hyperlinks
\usepackage{url}            % simple URL typesetting
\usepackage{booktabs}       % professional-quality tables
\usepackage{amsfonts}       % blackboard math symbols
\usepackage{nicefrac}       % compact symbols for 1/2, etc.
\usepackage{microtype}      % microtypography
\usepackage[table,dvipsnames]{xcolor}
\usepackage{comment}
\usepackage{wrapfig,lipsum,booktabs}

\usepackage{soul}
\usepackage{amsmath}
\usepackage{amssymb}
\usepackage{amsthm}
\usepackage{graphicx}
\usepackage{xspace}
\usepackage{multirow}
\usepackage{adjustbox}

\usepackage{algorithm}
\usepackage{algpseudocode}
\algrenewcommand\algorithmiccomment[1]{\hfill{\footnotesize$\triangleright$~#1}}

\usepackage{setspace} % added by Alp

\usepackage{graphicx} % added by Alp

\usepackage{arydshln} % added by Alp

\usepackage{etoc}

\newtheorem{theorem}{Theorem}[section]
\newtheorem{proposition}[theorem]{Proposition}

\newtheorem{corollary}[theorem]{Corollary}
\theoremstyle{definition}
\newtheorem{definition}[theorem]{Definition}

\theoremstyle{remark}

\newcommand{\methodname}{\textsc{KINDER}\xspace}

\title{\methodname: \underline{K}ernel-based \underline{Ind}ependence for Fair \\ Representation Learning via Prototype-space  \underline{Er}asure}

\author{%
  Abtin Mogharabin\textsuperscript{1,2}
  \quad
  Alp Toykan Kaplan\textsuperscript{3}
  \quad
  Jiaee Cheong\thanks{Equal senior authorship} ~\textsuperscript{4,5}
  \quad
  Sinan Kalkan\textsuperscript{$*$~3}
  \\[4pt]
  \textsuperscript{1}University of Tübingen
  \qquad
  \textsuperscript{2}Zuse School ELIZA
  \\
  \qquad
  \textsuperscript{3}Middle East Technical University
  \qquad
  \textsuperscript{4}Harvard University
  \qquad
  \textsuperscript{5}MIT
}

\begin{document}

\maketitle
\etocdepthtag.toc{main}
\begin{abstract}
%%%%%%%%%%%%%%%%%%%%%%%%%%%%%%%%%%%%%%
%%%%%%%%%%%%%%%%%%%%%%%%%%%%%%%%%%%%%%
%
%In the literature, 
%Contrastive learning  objectives have shown significant success across uni-modal and multi-modal tasks. 

%Achieving fairness in deep learning is difficult because protected information is often deeply entangled with the predictive structure in learned representations. Prior methods typically define debiasing indirectly through the failure of a chosen adversary, probe, or fairness penalty, making the target of bias concept erasure dependent on a particular decoder family or optimization setup. 
Concept erasure is a prominent approach to achieving fairness in machine learning through the removal of sensitive attributes from learned representations. 
Prior concept erasure methods typically define debiasing indirectly through the failure of a chosen adversary, probe, or fairness penalty, which makes the 
%\textcolor{teal}{
target of erasure dependent on a particular decoder family or optimization setup.
%}. 
%
%We introduce \methodname, a simple and scalable concept-erasure block that instead operates directly on representation geometry. 
%
%To address this challenge, we introduce \methodname, a simple, scalable, and model-agnostic block that can be inserted into deep learning pipelines to remove components of a representation that are statistically associated with protected attributes. 
%
% Rather than depending on a particular architecture, task type, or training paradigm, 
We introduce \methodname, which directly removes sensitive-attribute information from intermediate representations with respect to a chosen family of prototype functionals, and is compatible with any deep model containing an intermediate feature space, independent of the architecture, task type, or training paradigm.
%
% Past works in contrastive learning predominantly prioritize multi-view contrastive objectives while neglecting a mechanism to isolate task-irrelevant or sensitive variation beyond suppression via optimization penalties (e.g., MI bounds), which limits fairness control.
%
\methodname operates in a random Fourier feature space, where it estimates a sensitive subspace from protected group prototypes and projects representations onto its orthogonal complement. Thus, it forms an explicit representation cleaning mechanism distinct from prior fairness control studies.
We conduct extensive experiments across diverse settings and show the effectiveness of \methodname on (i) supervised and self-supervised settings, (ii) classification, regression, and image segmentation tasks, and (iii) diverse data modalities, including visual, textual, and tabular data. Code and trained models are available at: https://github.com/abtinmU/KINDER
%
%Our results show that \methodname significantly improves fairness across various protected attributes (i.e., gender, race, and age), while maintaining or improving the predictive performance.
%%

\end{abstract}

%%%%%%%%%%%%%%%%%%%%%%%%%%%%%%%%%%%%%%%%
\section{Introduction}

% Teaser layout controls

\begin{figure}
    \centering
    \includegraphics[width=0.98\linewidth]{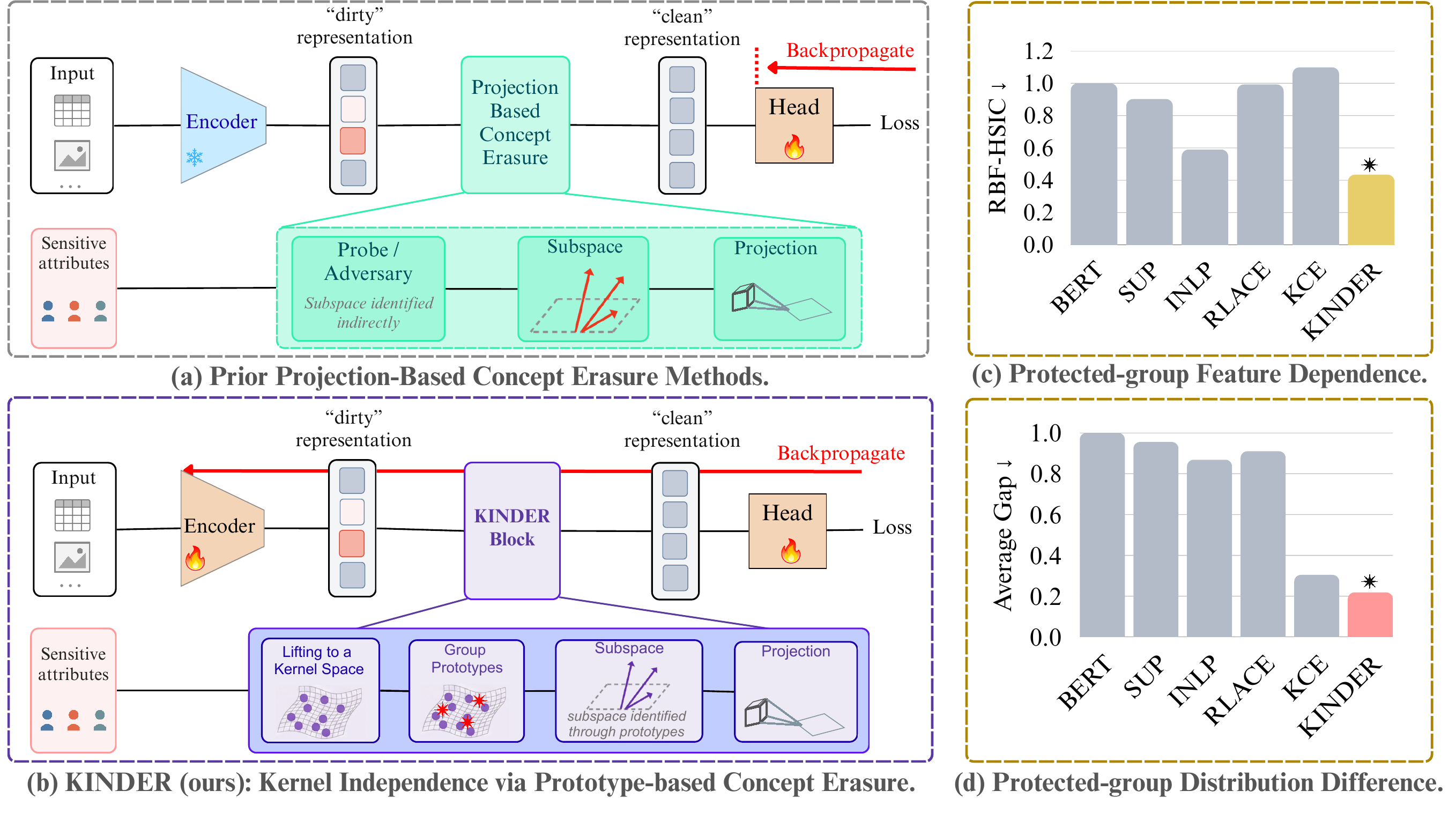}
    % \caption{Fundamental differences between (a) prior projection-based concept erasure methods and (b) \methodname. \methodname generalizes the concept erasure process via 
    % 1. lifting, 2. prototype estimation and 3. representation projection and cleaning which results in significantly lower 
    % \textcolor{red}{HSIC and average gap difference scores as demonstrated in (c) and (d)}
    % }
    \caption{Overview of prior projection-based concept erasure and \textbf{\methodname}. 
    (a) Prior methods identify an erasable subspace indirectly through decoder-specific objectives. 
    (b) \textbf{\methodname} instead builds protected-group prototypes in a lifted feature space and removes the directions along which those group prototypes differ.  
    (c--d) After applying \methodname on Moji, the representation should contain less protected-group information and smaller group-level distribution differences. This is shown by lower HSIC and average gap values. Significance tests are reported in Appendix~\ref{app:feature_level_bootstrap}.}
    \label{fig:teaser}
    \vspace{-7mm}
\end{figure}

% \textbf{Par1 1: Fair Representation Learning + concept erasusre}

%Extensive prior work in fair representation learning (FRL) has shown effectiveness in promoting fairness for specific downstream tasks \cite{luofair_neurips_205,liu2022fair,zemel2013learning}. 
%
%This is typically achieved via methods such as  ..., ...,  ... 
%
Extensive prior work in fair representation learning (FRL) has shown effectiveness in promoting fairness by learning intermediate representations that retain task-relevant information while reducing dependence on protected attributes \cite{zemel2013learning,edwards2015censoring,liu2022fair,luofair_neurips_205}. 
This can be pursued through adversarial invariance, information-theoretic objectives,
Maximum Mean Discrepancy (MMD) 
regularization, disentanglement, and certified downstream fairness guarantees \cite{madras2018learning,song2019learning,deka2023mmd,creager2019flexibly,jovanovic2023fare}. 
While a stream of FRL studies optimize the representations for a specific downstream task \cite{liu2022fair,cheong2024fairrefuse}, numerous FRL methods learn more \textit{general} representations that are fair \cite{zemel2013learning,chowdhury2022learning,priyadarshini2026learning,luofair_neurips_205}. 
This distinction is important because learned representations are often reused with unknown downstream tasks, prediction heads, modalities, or deployment settings, where a fairness constraint optimized for one task may not transfer to another \cite{madras2018learning,shen2023fair,song2019learning}. 
To achieve this reusable form of FRL, a substantial body of research focuses on learning representations that exclude sensitive concepts, such as demographic attributes like gender or race \cite{chowdhury2022learning,belrose2023leace,ravfogel2022adversarial}.
Such methods are known as \textit{concept erasure} methods \cite{lu2024mace,belrose2023leace}
and typically entail balancing two competing objectives: 
(i) removing concept-related information from the representations to ensure they are no longer predictive of the concept after erasure \cite{ravfogel2022adversarial,ravfogel2022linear}, 
and (ii) preserving as much information as possible from the original representation space \cite{saillenfest2025nonlinear,song2019learning,zhao2022inherent,chowdhury2025fundamental}.
%
\begin{comment}
One category of these methods draws inspiration from information
and probability theory [26, 31, 35, 37, 45, 54, 63, 65, 68, 72]. One work explores the use of
distance covariance [44]. 
%
Some methods can limit downstream unfairness by constraining the
total variation distance between the representation distributions of different groups [5, 48, 66, 76].
%
Other approaches promote independence from sensitive attributes through penalizing Maximum
Mean Discrepancy [15, 47, 56] or statistical dependence [25, 61], meta-learning [57], PCA [39, 42],
learning a shared feature space between groups [12], or disentanglement [14, 46, 55]. 
%
A stream of work uses adversarial training that limits the adversary’s performance in predicting sensitive
attributes [20, 22, 36, 60, 71].
%
However, ... 
\end{comment}
%
%%%%%%%%%%%%%%%%%%%%%%%%%%%%%%%%%%%%%%%%%%%%%%%%%
%
% \textbf{Part 2: Projection-based + definition of concept erasure + problem of model dependency}

A prominent family of concept erasure methods relies on projections \cite{ravfogel2020null,ravfogel2022linear,ravfogel2022adversarial,belrose2023leace,quirke2025slowing,shao2023gold,shi2024debiasing},
%the idea of projections. 
%
%Projection-based concept erasure methods 
and they typically
aim to identify a subspace associated with the protected concept 
\cite{ravfogel2020null,ravfogel2022linear,ravfogel2022adversarial}
and remove it from the representation through an explicit projection operator
%, which has made projection-based erasure a particularly influential direction in the concept-erasure literature 
\cite{belrose2023leace,quirke2025slowing,shao2023gold,shi2024debiasing} (Fig. \ref{fig:teaser}(a)). 
However, a key limitation is that the target of erasure is typically defined indirectly through restricted recovery families, such as linear probes or classifiers \cite{ravfogel2020null,belrose2023leace}, linear or kernelized adversarial games \cite{ravfogel2022linear,ravfogel2022adversarial}, quadratic decoders \cite{quirke2025slowing}, or other selected decoder families, rather than through independence at the level of the cleaned representation (\textbf{Observation 1 - O1}).
%
%\textbf{Research Gap 1 (RG1) (Decoder-relative erasure vs.\ representation-level erasure)}: can projection-based concept erasure be defined directly at the level of the cleaned representation, rather than through the failure of a selected decoder family?
%
A second limitation concerns \textit{how} the erasable subspace %itself 
is identified. 
Existing projection-based methods typically recover the removable directions indirectly, through probe-based optimization \cite{ravfogel2020null}, minmax or adversarial objectives \cite{ravfogel2022linear,ravfogel2022adversarial,chowdhury2021adversarial}, covariance-based closed forms \cite{belrose2023leace}, sufficient or spectral projections \cite{shi2024debiasing,shao2023gold}, or kernel/rate-distortion objectives \cite{basu2023robust}.
%} \cite{ravfogel2020null,ravfogel2022linear,ravfogel2022adversarial,belrose2023leace,shao2023gold,shi2024debiasing}.
%
This leaves open whether group-conditioned nuisance structure can instead be estimated directly from the representation geometry itself, rather than inferred through the selected decoder family used to test erasure (\textbf{O2}). 
%
%At the same time, 
Moreover,
concept erasure methods %that work on the idea of projections 
are %most naturally 
often applied as post-hoc transforms on fixed representations (\textbf{O3}) \cite{ravfogel2020null,ravfogel2022linear,belrose2023leace,shao2023gold}.  
However, in supervised settings, once the representation is projected, the original task head is no longer matched to the cleaned features, and %the 
post-hoc projections can alter the statistical geometry of the feature distribution \cite{johansson2024happens}. 
This impedes direct comparison to native end-to-end models and can introduce an additional utility bottleneck \cite{chowdhury2025fundamental,zhao2022inherent}.
Further, realistic fairness settings often require simultaneous control of multiple protected attributes \cite{tian2024multifair,pessach2022review}, 
including intersectional or structured subgroup settings where marginal fairness can hide subgroup violations \cite{kearns2018preventing,foulds2020intersectional,creager2019flexibly,vethman2025fairness}, 
and imbalanced groups \cite{jones2025examining}. 
Yet many concept erasure methods are predominantly evaluated in single-attribute regimes \cite{ravfogel2020null,ravfogel2022linear,ravfogel2022adversarial,belrose2023leace,shao2023gold},
thus calling into question whether they remain stable and effective when multiple protected attributes are removed simultaneously \cite{shi2024debiasing,yang2025fairsmoe} (\textbf{O4}).  

Based on observations O1-O4, we address three main research questions: 
\textbf{Research Question 1 (RQ1):} Can a projection-based operator estimate erasable group-conditioned structure directly at the representation level? 
\textbf{RQ2}: Can projection-based concept erasure be defined at the cleaned-representation level, rather than through the failure of a selected decoder family?
\textbf{RQ3:} Can such an operator remain useful across post-hoc and end-to-end training, diverse task types, and single, multiple, intersectional, or imbalanced protected-attribute settings?
%
%\textbf{RG4}: Can a single projection-based concept erasure operator remain effective under multiple, intersectional, and imbalanced protected-attribute settings?

\begin{table*}[hbt!]
\addtolength{\tabcolsep}{-1mm}
\caption{Prior methods guarantee erasure only for restricted decoder families. \methodname instead proves a representation-level guarantee, $R\perp_{\mathfrak F} S$, meaning independence (invariance) up to the degree captured by the chosen prototype space ${\mathfrak F}$.  
The all-$f$ guarantee assumes the affine model (Theorem~\ref{thm:canonical_affine_projection}).
Thus, under \methodname, removal works at the level of the cleaned representation itself.
Further, \methodname has been evaluated on a wider set of data, tasks and settings.
Abbreviations:
\textbf{C}: classification.
\textbf{R}: regression.
\textbf{S}: segmentation.
\textbf{Set.}: settings.
\textbf{Inter.}: intersectional fairness.
\textbf{E}: end-to-end.
\textbf{P}: post-hoc.
\textbf{NR}: Not Reported.
\textbf{Rep.}: Reported.
\textbf{T}: Text. 
\textbf{Tab}: Tabular.
\textbf{V}: Visual.
  }   \label{tab:comparative_summary}
  \centering
  \scriptsize
  \begin{tabular}{l|l|ll|l|c|c|c}
    \hline
    \textbf{Method} & \textbf{Independence type} & \textbf{Erasure target} & \textbf{Subspace source} 
  &\textbf{Data} &\textbf{Task} &\textbf{Set.} & \textbf{Inter.} \\  
%     \hline
    %\cdashline{3-12}[.4pt/2pt]%(r){2-14}   
%    \hspace*{1cm}
 %  && &  \\

   \hline
 
INLP 
\cite{ravfogel2020null}
&
$f(R)\perp S$, $\forall$  \textit{linear $f$} 
&
Linear probe failure &
Iterative probes; nullspace  &T &C &P &NR\\

RLACE 
\cite{ravfogel2022linear}
&
$f(R)\perp S$, $\forall$  \textit{linear $f$} 
&
Linear adversary failure &
Linear minmax subspace  &T &C &P &NR\\

KCE 
\cite{ravfogel2022adversarial}
&
$f(R)\perp S$, for some \textit{kernel $f$} 
%\textit{chosen kernel $f$} 
&
Kernel adversary failure &
Kernelized minmax subspace  &T &C &P &NR\\

SAL 
\cite{shao2023gold} %, shao2023erasure}
&
$\mathrm{Cov}(f(R),g(S)) \approx 0$, for
&
High-covariance directions &
Covariance SVD projection  &T &C &P &NR\\

&top \textit{spectral linear} pairs $(f,g)$ &  & & && \\

LEACE
\cite{belrose2023leace}
&
$f(R)\perp S$, $\forall$  \textit{linear $f$} 
&
Linear classifier failure &
Covariance projection  &T &C &P &NR\\

SUP 
\cite{shi2024debiasing}
&
$f(R)\perp S$ for some \textit{chosen $f$} 
&
Unbiased subspace &
Sufficient projection &T &C &P &NR\\

QLEACE  
\cite{quirke2025slowing}
&
$f(R)\perp S$, $\forall$ \textit{quadratic $f$} 
&
Quad. decoder failure &
Closed-form quad. erasure &T &C &P &NR\\

\midrule

 \rowcolor{gray!10} \textbf{\methodname} &
Pop. guarantee (affine model): & Prototype-space invariance &
Prototype SVD projection &T  &C &E &Rep.\\  
\rowcolor{gray!10}
\textbf{(ours)} &Theorem: $R \perp_{\mathfrak{F}} S$ $\Rightarrow $ Cor.: &&&Tab & R & P &\\

\rowcolor{gray!10}
& $f(R)\perp_{\mathfrak F} S$ \textit{\textbf{$\forall$ measurable}} \textbf{$f$}  &&& V& S &&\\

\bottomrule

    \hline
  \end{tabular}
%  \vspace*{-0.2cm}
  
\vspace*{-0.3cm}
\end{table*}

To address these gaps, we introduce \methodname (see Fig. \ref{fig:teaser}), 
a modular projection-based concept erasure operator that 
(i) lifts representations into a prototype space, 
(ii) estimates group-conditioned nuisance structure from explicit group prototypes, and 
(iii) removes that structure through an orthogonal projection. 
Unlike prior projection-based methods whose guarantees are tied to linear probes or minmax games \cite{ravfogel2020null,ravfogel2022linear}, kernelized adversaries \cite{ravfogel2022adversarial}, closed-form linear or quadratic erasure \cite{belrose2023leace,quirke2025slowing}, sufficient or spectral projections \cite{shi2024debiasing,shao2023gold}, or iterative RKHS functional erasure \cite{akbari2026obliviator}, 
\methodname defines erasure directly at the level of the cleaned representation itself. 
This makes \methodname decoder-agnostic and model-agnostic, while preserving the explicit and interpretable structure of projection-based concept erasure. 

Our key contributions (KC) are as follows. 
\textbf{KC1 (addresses RQ1 and RQ2):} 
%
%We reformulate projection-based concept erasure from decoder-relative non-recoverability to representation-level erasure 
%
We improve projection-based concept erasure generalizability
by moving the target of removal from a restricted decoder family to the cleaned representation itself. 
\textbf{KC2 (addresses RQ2 and RQ3):} We introduce a prototype-space projection mechanism that estimates erasable group-conditioned structure directly from explicit group prototypes in a nonlinear lifted space and 
%rather than inferring it indirectly from probes, adversaries, or decoder-dependent procedures. 
%
%By storing explicit prototypes for each protected group, \methodname keeps persistent group-conditioned statistics over training, 
makes it easier to recover the shared nuisance structure 
%when the 
across  
\textit{multiple}, \textit{intersectional}, or \textit{unevenly represented groups}.
\textbf{KC3 (addresses RQ3):} %We show that the same 
\methodname %operator 
supports both end-to-end training and retrofit adaptation within a unified formulation. %.avoiding the clean separation between training-time fairness and post-hoc concept erasure that characterizes much prior work. 
\textbf{KC4 (addresses RQ1--RQ2):} We provide a population-level theory, summarized in Section~\ref{subsec:theory}, showing that \methodname removes group-specific nuisance perturbations in prototype space and reaches prototype-space independence of the cleaned representation from the protected attributes. 
\vspace{-2mm}
\section{Related Work}
\vspace{-2mm}

%%%%%%%%%%%%%%%%%%%%%%%%%%%%%%%%%%%

%\subsection{Concept Erasure Methods}

%Concept erasure is relevant to numerous applications, including bias mitigation for fairness \cite{belrose2023leace,ravfogel2022linear} and improving explainability \cite{abraham2022cebab}
%
%Existing concept erasure approaches can be categorized as post-hoc and fine-tuned.
%
The simplest setting for concept erasure involves categorical
concepts, % which encompass a wide range of applications, 
such as the removal of demographic biases related to gender or race in text
embeddings \cite{bolukbasi2016man,chowdhury2022learning,basu2023robust}
and can be achieved by both
adversarial training 
\cite{edwards2015censoring,chowdhury2021adversarial,wang2021dynamically,li2018towards}
or projection-based methods \cite{bolukbasi2016man,gonen2019lipstick,ravfogel2020null,ravfogel2022linear,ravfogel2022adversarial,shao2023gold,chowdhury2021adversarial,basu2023robust}.
Post-hoc methods directly modify the learned representations without altering the parameters of the backbone model \cite{ravfogel2020null,akbari2026obliviator}. 
In contrast, fine-tuned approaches require access to the model and enforce erasure by finetuning its parameters \cite{shao2023gold}. 
%
%A summary of concept-erasure methods can be found in Table \ref{}.
%
%
Earlier projection-based methods typically rely on linear projection techniques~\cite{bolukbasi2016man, gonen2019lipstick, ravfogel2020null, ravfogel2022linear} applied in a post-hoc manner, whereas recent work has attempted to extend them to non-linear settings~\cite{basu2023robust, chowdhury2021adversarial, ravfogel2022adversarial, shao2023gold}.

\textbf{Comparative Summary:}
Table~\ref{tab:comparative_summary} compares \methodname against projection-based concept-erasure methods. 
KCE extends minimax erasure to a selected kernel family \cite{ravfogel2022adversarial}; 
SAL removes guarded information through spectral covariance directions \cite{shao2023gold}; 
LEACE and QLEACE provide closed-form guarantees for linear and quadratic information \cite{belrose2023leace,quirke2025slowing}; 
and SUP removes information through sufficient projection \cite{shi2024debiasing}. 
Thus, within projection-based concept erasure, existing guarantees are typically tied to restricted decoder families, post-hoc editing regimes, or limited protected-attribute evaluations. \methodname instead proves a more general, decoder-agnostic representation-level guarantee. %rather than a unified representation-level projection operator supporting both end-to-end and retrofit use across single, multiple, and intersectional protected-attribute settings.
%
% Certified FRL methods such as FARE provide complementary guarantees on downstream classifier unfairness, but their guarantees concern downstream prediction fairness rather than representation-level removal of protected-attribute information which \methodname targets \cite{jovanovic2023fare}.

% Kernel and RKHS tools have also been used for fair representation learning and prediction debiasing, including fair RKHS subspaces for kernel models \cite{tan2020learning}, MMD-based two-sample indistinguishability between sensitive groups \cite{deka2023mmd}, and RKHS-based debiasing of CLIP zero-shot predictions \cite{dehdashtianfairerclip}. 
% %
% These methods show that kernel machinery is useful for fairness, but they operate as model-aware fair representations, statistical tests/objectives, or prediction-level debiasing procedures rather than erasure operators. 
% %
% \methodname uses the RKHS space for a different purpose of estimating group-conditioned nuisance directions from protected-group prototypes and the removing those directions.
% Orthogonal to this comparison, one recent work studies information-theoretic limits of perfect concept erasure and shows that erasure can inherently trade off with representation utility \cite{chowdhury2025fundamental}. \methodname does not bypass these limits, but targets a different limitation of projection-based erasure, namely the decoder-family dependence.

Kernel and Reproducing kernel Hilbert space
(RKHS) tools have also been used for fair representation learning and prediction debiasing, including fair subspaces \cite{tan2020learning}, MMD-based objectives \cite{deka2023mmd}, and CLIP debiasing \cite{dehdashtianfairerclip}. 
These methods show that kernel machinery is useful for fairness, but focus on statistical tests/objectives or prediction-level debiasing, whereas 
\methodname uses RKHS/
Random Fourier Features (RFF) prototypes to estimate and project out group-conditioned nuisance directions. 
Orthogonal to this comparison, recent work studies information-theoretic erasure--utility limits of perfect concept erasure \cite{chowdhury2025fundamental}. 
\methodname does not bypass these limits, but targets a different limitation of projection-based erasure, namely, the decoder-family dependence \cite{belrose2023leace,quirke2025slowing}.

\begin{figure}[ht]
\vskip 0.2in
\begin{center}
\centerline{\includegraphics[width=0.99\columnwidth]{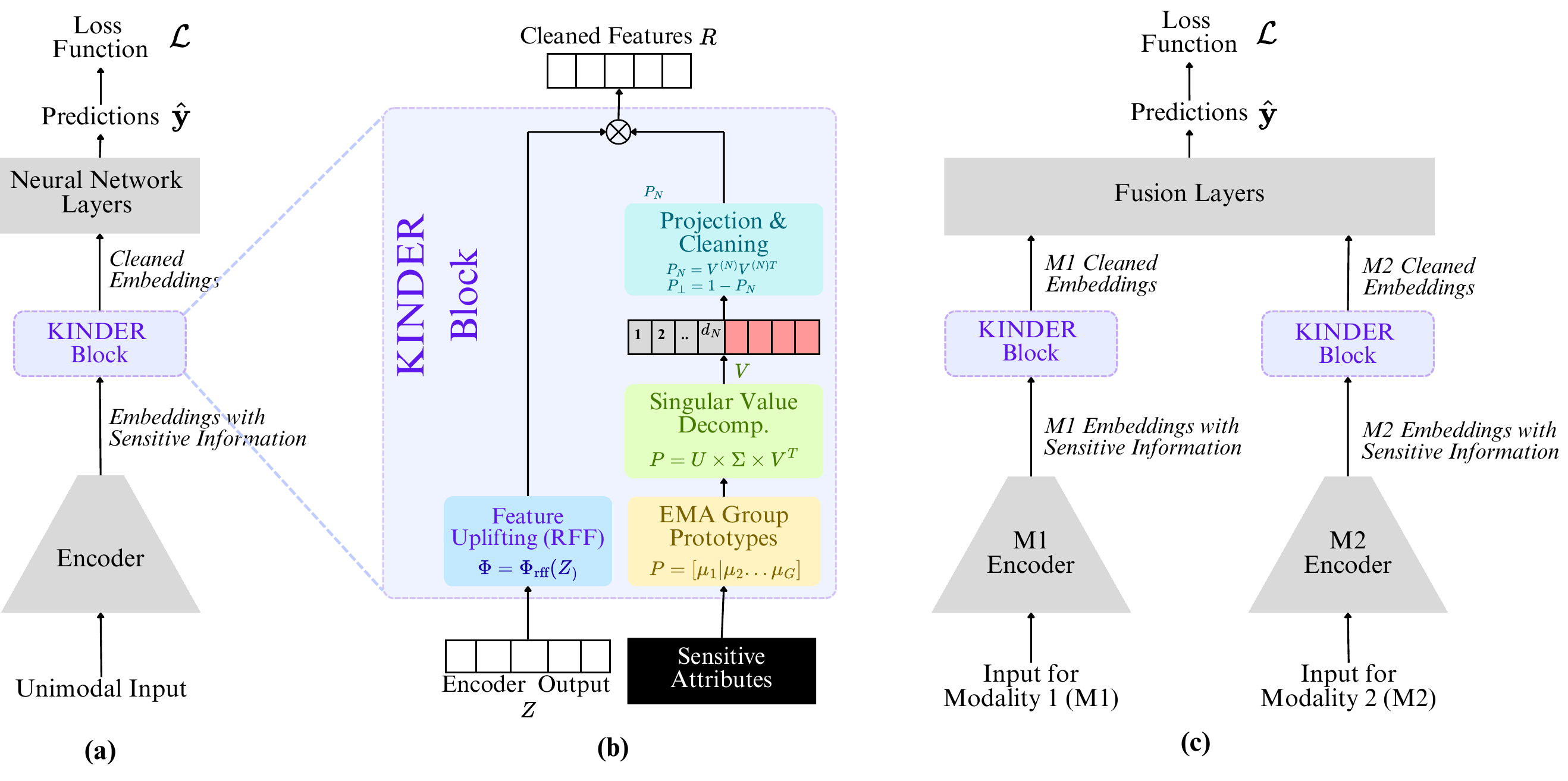}}
\vspace{-0.3cm}
\caption{\textbf{The \methodname block.} 
\textbf{(a) Unimodal use: \methodname} is inserted between the encoder and the downstream network to transform embeddings containing sensitive variation into cleaned embeddings used by the task loss. 
\textbf{(b) Inside the block}, the encoder output \(Z\) is first lifted into a prototype space through random Fourier features (RFF); group-conditioned prototypes are then maintained by exponential moving averages (EMA) using the sensitive attributes, and their dominant variation is summarized by a singular value decomposition (SVD) to estimate the nuisance subspace \(N\). The cleaned representation is obtained by projecting onto its orthogonal complement, \(P_{\perp}=I-P_N\), while the removed component corresponds to the estimated group-associated nuisance structure. 
\textbf{(c) Multimodal use:} the same block is applied independently to each modality branch before fusion, so downstream layers operate on cleaned modality-specific embeddings.}\label{fig:kinder_overview}
\end{center}
\vskip -0.5cm
\end{figure}
% RFF lift => switch to HS
% Ra, Rb => directly provided to SSL loss.

\vspace{-1mm}
\section{Preliminaries and Background}
\label{subsec:prelim}
\vspace{-1mm}

\paragraph{Notation and Problem Setting.}
% Without loss of generality, we assume a bi-modal setting with paired inputs \((x_{m_1} \in X_{m_1},x_{m_2} \in X_{m_2})\) from two modalities (${m_1}$ and ${m_2}$)  with encoders \(f_{m_1},f_{m_2}\) providing %us 
% their respective feature representations:
% \begin{eqnarray}
% z_{m_1}=f_{m_1}(x_{m_1}; \theta_{m_1})\in Z_{m_1} \subset \mathbb{R}^{d},\\ 
% z_{m_2}=f_{m_2}(x_{m_2}; \theta_{m_2})\in Z_{m_2} \subset \mathbb{R}^{d},
% \end{eqnarray}
% with $\theta_{m}$ denoting the parameters of the encoder $f_{m}$ for $m\in \{m_1, m_2\}$. 
% % We assume a supervised classification setting where we have a single target label for a paired input: $((x_{m_1},x_{m_2}), y)$.

% Let \(S\) be a discrete sensitive attribute with values in \(\{1,\ldots,G\}\).
% For sample \(i\), we write \(s^{(i)}\in\{1,\ldots,G\}\) for the value of \(S\).

% We use mini-batches \(\{(x_{m_1}^{(i)},x_{m_2}^{(i)},s^{(i)})\}_{i=1}^{B}\). When task supervision is available, we additionally write \(y^{(i)}\) for the corresponding target.
Let \(x\in\mathcal X\) be an input, \(f_\theta\) an encoder, and \(z=f_\theta(x)\in\mathbb R^d\) its representation; for two-view or multimodal settings, the same notation is applied to each branch. 
Let \(S\in\{1,\ldots,G\}\) be a discrete sensitive attribute, with \(s^{(i)}\) denoting the protected-group label for sample \(i\). 
We use mini-batches \(\{(x^{(i)},s^{(i)})\}_{i=1}^{B}\), and add \(y^{(i)}\) when task supervision is available. 
For a batch of features \(\mathbf Z\in\mathbb R^{B\times d}\), we define the centering matrix \(\mathbf H_B=\mathbf I_B-\frac{1}{B}\mathbf 1\mathbf 1^\top\). We define \(\mathbf Z_c=\mathbf H_B\mathbf Z\) for the column centered version of \(\mathbf Z\).

\vspace{-2mm}
\paragraph{Reproducing Kernel Hilbert Space (RKHS).}
A Hilbert space \(\mathcal H\) is a vector space %equipped 
with an inner product \(\langle \cdot,\cdot\rangle_{\mathcal H}\), which allows geometric notions such as distance and orthogonality to be defined in possibly infinite-dimensional spaces. An RKHS is a Hilbert space of functions where evaluation is continuous and satisfies
$
f(x)=\langle f,\,k(\cdot,x)\rangle_{\mathcal H},
$
where \(k\) is the reproducing kernel.
In our setting, bounded measurable kernels \(k_Z\) and \(k_S\) induce RKHSs \(\mathcal H_Z\) and \(\mathcal H_S\), with feature maps
$\varphi_Z(z)=k_Z(\cdot,z)$ and $\varphi_S(s)=k_S(\cdot,s)$
for encoder features \(z\) and sensitive attributes \(s\). 
This lifts \(Z\) and \(S\) into Hilbert spaces where \textit{distributional differences can be expressed through inner products}.
\vspace{-1mm}
\section{\methodname: Methodology}
\vspace{-1mm}

%Our goal is to transform intermediate representations produced by a model into ``cleaned'' representations \(R\) that preserve task-relevant information while removing dependence on \(S\) arising from the group-specific nuisance structure.
%
%
Intuitively, our \methodname block removes group-specific nuisance structure in the lifted prototype space and maps each encoder output to a ``cleaned'' representation retaining only the shared component of the representation (see Figure~\ref{fig:kinder_overview} -- see Appendix \ref{sect:appendix_KINDER_algorithm} for the full algorithm). We first introduce \methodname for a unimodal setting and extend it for a pair of representations \((z_a,z_b)\) (corresponding to e.g., two views, augmentations, or modalities). 
%, but the block applies %identically  to single representations as well. 
\methodname executes three steps:
\vspace{-2mm}
\begin{enumerate}
    \item Step 1: Lifting the encoder output into an RKHS space (Section \ref{sect:method_lifting}).
    \item Step 2: Prototype Estimation in the RKHS space (Section \ref{sect:method_prototype}).
    \item Step 3: Representation projection and cleaning in the RKHS space (Section \ref{sect:method_projection}).
\end{enumerate}
In Section~\ref{subsec:theory}, we provide a theoretical analysis showing that, under the corresponding population-level structural model, the cleaned representation becomes independent of the protected attribute at the prototype-space level, which is certified by zero pairwise MMD across groups.

\vspace{-2mm}

\subsection{Step 1: Uplifting the Encoder Features into an RKHS}
\label{sect:method_lifting}

%
%We now describe how \methodname lifts encoder features into a prototype space and constructs the removable nuisance subspace in that space. 
%

\paragraph{Kernels on \(Z\) and \(S\).}
On encoder features $Z$, we use the Gaussian RBF kernel with bandwidth \(\gamma>0\):
\begin{equation}
k_Z(z,z') \;=\; \exp\!\Big(-\tfrac{\|z-z'\|_2^2}{2\gamma^2}\Big)
\;=\; \langle \varphi_Z(z),\,\varphi_Z(z')\rangle_{\mathcal H_Z},
\end{equation}
which is a standard choice because it is universal and characteristic on \(\mathbb{R}^d\), allowing the lifted representation to capture rich geometric differences between group-conditional distributions in the encoder feature space \cite{muandet2017kernel,sriperumbudur2011universality}.
On the sensitive attribute, we use the Dirac kernel:
$
k_S(s,s') \;=\; \mathbf{1}\{s=s'\},$ where $\varphi_S(s)=e_s\in\mathbb{R}^{G}.
$

For discrete sensitive attributes, this kernel coincides with the one-hot embedding and makes the RKHS mean  \(\mu_S=\sum_s \pi_s e_s\), with \(\pi_s=\Pr(S=s)\), and the class conditional means \(\mathbb E[\varphi_Z(z)\mid S=s]\) directly interpretable as group prototypes \cite{muandet2017kernel}.
This also makes group-conditioned prototype statistics directly interpretable in the lifted space, which is the basis for the nuisance-subspace construction used by \methodname.

\paragraph{Random Fourier features for \(k_Z\).}

We approximate \(k_Z\) with random Fourier features (RFF) \citep{rahimi2007random}. For bandwidth \(\gamma>0\) and RFF dimension \(M\),
\begin{equation}\footnotesize
\Phi_{\mathrm{rff}}(Z)
=\sqrt{\tfrac{2}{M}}\,
\big[\cos(\omega_1^\top Z+b_1),\ldots,\cos(\omega_M^\top Z+b_M)\big]^\top
\in\mathbb{R}^{M},
\label{eq:rff}
\end{equation}
with \(\omega_j\overset{\mathrm{iid}}{\sim}\mathcal N(0,\gamma^{-2}I_d)\) and \(b_j\overset{\mathrm{iid}}{\sim}\mathrm{Unif}[0,2\pi]\), where \(I_d\) is the \(d\times d\) identity matrix.
For notational simplicity we write \(\Phi\) for \(\Phi_{\mathrm{rff}}\) in what follows. All linear operators used in the \methodname block implementation (prototypes, projectors, and cleaned features) act in this lifted space \(\mathbb{R}^{M}\).

\subsection{Step 2: Prototype Estimator of the Sensitive Subspace via SVD}
\label{sect:method_prototype}

Let \(X=\Phi_{\mathrm{rff}}(z)\in\mathbb{R}^{M}\) denote the lifted encoder feature. 
Rather than fixing the removable structure to a single statistic, \methodname estimates \textit{group-associated variation} through a chosen prototype family. 
For an order parameter \(K\), let
$
\mathcal F_K=\{\psi^{(q)}_{a}: q=1,\ldots,K,\ a\in\mathcal A_q\}
$
be a finite collection of prototype functionals, where each \(\psi^{(q)}_{a}\) maps lifted features to a vector in \(\mathbb{R}^{M}\). 
For each group \(s\), order \(q\), and prototype index \(a\), we maintain an EMA prototype
\begin{equation}
\begin{aligned}
p^{(q)}_{s,a}
&\leftarrow
(1-\rho)\,p^{(q)}_{s,a}
+
\rho\cdot
\frac{1}{|\mathcal B_s|}
\sum_{i\in \mathcal B_s}
\psi^{(q)}_{a}\!\left(X^{(i)};\mu_s\right),\\
\mathcal B_s &:= \{\,i:\ s^{(i)}=s\,\}.
\end{aligned}
\label{eq:general_proto}
\end{equation}
Here, \(\rho\in(0,1]\) controls the EMA update rate, and \(\mu_s=\mathbb E[X\mid S=s]\) is the group-conditional mean used for central higher-order moments; for first-order prototypes, this centering argument is not needed.

For each prototype statistic, we define its empirical global average
$
\bar p^{(q)}_{a}
=
\sum_{s=1}^{G}\hat\pi_s\,p^{(q)}_{s,a},
$
where \(\hat\pi_s\) is the empirical group frequency. 
The order-\(K\) nuisance subspace is then defined as the span of all centered group-prototype differences:
\begin{equation}
\mathcal N_K
:=
\mathrm{span}
\Big\{
p^{(q)}_{s,a}-\bar p^{(q)}_{a}
:
s=1,\ldots,G,\ q=1,\ldots,K,\ a\in\mathcal A_q
\Big\}
\subset \mathbb{R}^{M}.
\label{eq:general_nuisance_subspace}
\end{equation}

To estimate this subspace, \methodname processes the prototype orders sequentially. 
For each order \(q\), we form a centered prototype matrix \(\tilde P_q\) by stacking the rows
$
(p^{(q)}_{s,a}-\bar p^{(q)}_{a})^\top
$
over all groups \(s\) and prototype indices \(a\in\mathcal A_q\).
For the first order, we directly compute the thin SVD of \(\tilde P_1\). 
For higher orders, before applying SVD, we remove the directions already selected from lower-order prototypes:
\begin{equation}
\tilde P_{q\mid <q}
=
\tilde P_q
\left(I_M-Q_{<q}Q_{<q}^{\top}\right),
\end{equation}
where \(Q_{<q}\) is the orthonormal basis selected from orders \(1,\ldots,q-1\). 
We then compute
\begin{equation}
\tilde P_{q\mid <q}
=
U^{(q)}\Sigma^{(q)}V^{(q)\top},
\label{eq:proto_svd}
\end{equation}
and add the leading right singular vectors to the nuisance basis.
After processing orders \(1,\ldots,K\), let \(Q_{\le K}\) denote the resulting basis. 
We define
\begin{equation}
P_N^{(K)} = Q_{\le K}Q_{\le K}^{\top},
\qquad
P_\perp^{(K)} = I_M-P_N^{(K)}.
\label{eq:proj_modality}
\end{equation}
The selected singular vectors span the retained nuisance subspace, which coincides with \(\mathcal N_K\) when all nonzero directions are retained and otherwise provides a rank-truncated approximation. Higher-order prototypes can contribute additional nuisance directions, while the cleaning operation remains a single orthogonal projection.

For first-order \methodname, group means in the RFF space approximate kernel mean embeddings, so their differences capture nonlinear group-conditioned variation beyond ordinary feature-mean differences. Since linear directions in this lifted space correspond to nonlinear functions of the original encoder features, the SVD projection can remove the selected directions of these discrepancies using linear operations in the lifted space.

\paragraph{Prototype-order instantiations.}
The simplest instantiation is first-order \methodname, with \(\psi^{(1)}(X)=X\). At the population level, the corresponding nuisance subspace is
\begin{equation}
\mathcal N_1
=
\mathrm{span}\{\mu_s-\bar\mu:\ s=1,\ldots,G\},
\qquad
\bar\mu=\sum_{s=1}^{G}\pi_s\mu_s.
\end{equation}
The EMA prototypes in Eq.~\eqref{eq:general_proto} provide the corresponding empirical estimates.
Higher-order variants address residual nuisance directions through additional prototype functionals while using the same projection pipeline. Second-order prototypes capture group covariance differences, while orders $q\ge 3$ use directional moment sketches to avoid constructing full higher-order tensors (Appendix~\ref{app:higher_order_prototypes}). A feature-level ablation over higher-order prototype families is also provided in Appendix~\ref{app:prototype_order_ablation}.

\subsection{Step 3: Representation Projection and Cleaning}
\label{sect:method_projection}

Given a batch of lifted features \(\mathbf X=\Phi_\mathrm{rff}(\mathbf Z)\in\mathbb{R}^{B\times M}\), we define the cleaned representations as
\begin{equation}
\mathbf R^{(K)} = \mathbf X\,P_{\perp}^{(K)}\in\mathbb{R}^{B\times M}.
\label{eq:clean_repr}
\end{equation}

Only \(\mathbf R^{(K)}\) is passed to the remainder of the model (e.g., a projection head, classifier, or regressor). Protected labels are used to estimate and update the nuisance projector. Once the projector is fixed, the same cleaning map is applied to all representations without requiring their individual protected labels.
For the theoretical analysis, \(R\) denotes the cleaned representation of a single branch, with any view or modality index suppressed.

\subsection{Training Objective (Loss)}
\label{subsubsec:loss}

\methodname is loss-agnostic: after cleaning, the representation is passed to the downstream head or objective without modifying the task loss. 
For a supervised input $x$ with target $y$, an encoder $f_\theta$ first produces
$
    z = f_\theta(x) \in \mathbb{R}^{d}.
$
\methodname then lifts and cleans this feature,
\begin{equation}
    R = \methodname(z,s) = \Phi_{\mathrm{rff}}(z) P_{\perp}^{(K)} \in \mathbb{R}^{M},
\end{equation}
and the downstream head $h_\omega$ is trained with the original supervised objective,
$
    \mathcal{L}_{\mathrm{sup}}^{\methodname}
    =
    \mathcal{L}_{\mathrm{task}}\!\left(h_\omega(R), y\right).
$
Thus, \methodname can be used with standard classification, regression, or segmentation losses without changing the task objective. 
For two-view self-supervised settings, the same block is applied to each view or modality before the contrastive objective; explicit uni-/multimodal SSL formulations are provided in Appendix~\ref{app:loss_formulations}.

\begin{comment}

\hl{Moved to appendix}
In supervised settings, the same cleaned representations \(R_m\) can be fed %to a task head 
\(h\) and trained with a standard loss (e.g., cross-entropy %for classification 
or regression loss) without modifying the \methodname block. 

%We also experiment on semi-supervised settings, where we use a two-stage protocol.
We use a two-stage protocol on semi-supervised settings.
%
In Stage~1 (self-supervised pretraining), we train on all available samples using either the unimodal objective \(\mathcal L_{\methodname}^{\mathrm{uni\text{-}ssl}}\) or the multimodal objective \(\mathcal L_{\methodname}^{\mathrm{mm\text{-}ssl}}\), while maintaining prototypes indexed only by \(s\) in Eq.~\eqref{eq:ema_proto}.
%
In Stage~2 (supervised fine-tuning), we train a task head on the labeled subset using the supervised objective \(\mathcal L_{\methodname}^{\mathrm{sup}}\).
%
When discrete task labels are available, Stage~2 also admits a label-aware refinement in which prototypes are indexed by \((s,y)\), i.e., by replacing \(s^{(i)}\) with \((s^{(i)},y^{(i)})\) in Eq.~\eqref{eq:ema_proto}. This changes only the prototypes used to form \(P_{\perp}\) and leaves the training objective unchanged. The resulting label-aware version conditions the removable nuisance structure on the task labels, reducing the risk of removing task-relevant variation in supervised classification settings.
%
\end{comment}

%
%
%\hl{moved the following to the appendix}

%%%%%%%%%%%%%%%%%%%%%%%%%%%%%%%%%%%%%%%%%%%%%%%%%%%%%%%%%%%%%%%%%%

\subsection{Theoretical Analysis}\label{subsec:theory}

%At a high level, \methodname operates in a chosen prototype space, estimates the group-conditioned nuisance structure induced by a selected family of prototype statistics, and removes it by orthogonal projection. 
% %
%The resulting guarantees therefore depend on the prototype family used to define that nuisance structure, rather than on a particular decoder or adversary family. 

In Appendix \ref{sect:appendix_theory}, we provide rigorous theoretical guarantees for \methodname, 
%demonstrating its capacity to achieve exact independence from a discrete protected attribute $S$. 
%
%We ground our 
beginning with a shared-base affine structural model. This model assumes that a generic representation $Z$ is generated by applying group-specific affine transformations to a latent vector $W$ that is strictly independent of $S$ ($W \perp S$). 
Under this model, we identify the canonical affine discrepancy subspace, $\mathcal{N}_\mathrm{aff}$, which captures the precise linear directions along which group-specific affine images of the same latent point diverge. As shown in Theorem \ref{thm:canonical_affine_projection}, projecting $Z$ onto the orthogonal complement of $\mathcal{N}_\mathrm{aff}$ yields a cleaned representation $R^*$ that is strictly independent of $S$. 
%Crucially, 
We prove %that 
$\mathcal{N}_\mathrm{aff}$ is the minimal linear subspace whose removal equalizes all groups, thereby preserving maximum latent information.

\methodname extends this affine intuition into a reproducing kernel Hilbert space (RKHS) via prototype families. By mapping representations to 
a Hilbert space $\mathbb{R}^{M}$ using $\Phi_{\mathrm{rff}}(Z)$, \methodname forms a cleaned representation $R$ by projecting away 
the subspace $\mathcal N_K$ spanned by the differences in group prototype means. 
Theorem \ref{thm:kinder_f_independence} establishes that this projection achieves prototype-level independence, 
%meaning that 
i.e. the cleaned representation is invariant to the protected attribute under the linear kernel of the prototype family ($R \perp_{\mathfrak{F}} S$). 
Further, Corollary~\ref{cor:exact_independence} guarantees that if the prototype functionals are rich enough to identify the relevant group laws in the cleaned representation space, zero MMD implies full probabilistic independence, $R \perp S$. 
The scope of this independence is determined by the selected prototype family: The projection equalizes its group-conditioned prototype statistics instead of asserting that all possible dependence is removed. Although our experiments use moment-order functionals, the framework also permits richer prototype families, including non-moment functionals, to control a broader class of group-conditioned variation without changing the projection mechanism. Complete proofs for all statements, including the minimality of the discrepancy subspace, are provided in Appendix~\ref{sect:appendix_theory}. 

In practice, RFF provides a finite-dimensional approximation of the Gaussian kernel, EMA estimates group-conditioned prototypes, and sequential SVD estimates the corresponding nuisance subspace. Proposition~\ref{prop:finite_sample_prototype} further bounds the residual prototype discrepancy from rank truncation and projector estimation, while Proposition~\ref{prop:rff_approximation} quantifies finite-RFF approximation of the Gaussian kernel. These results show the approximation errors beyond the population-level guarantee.

\begin{figure*}[t]
    \centering
    \includegraphics[width=\textwidth]{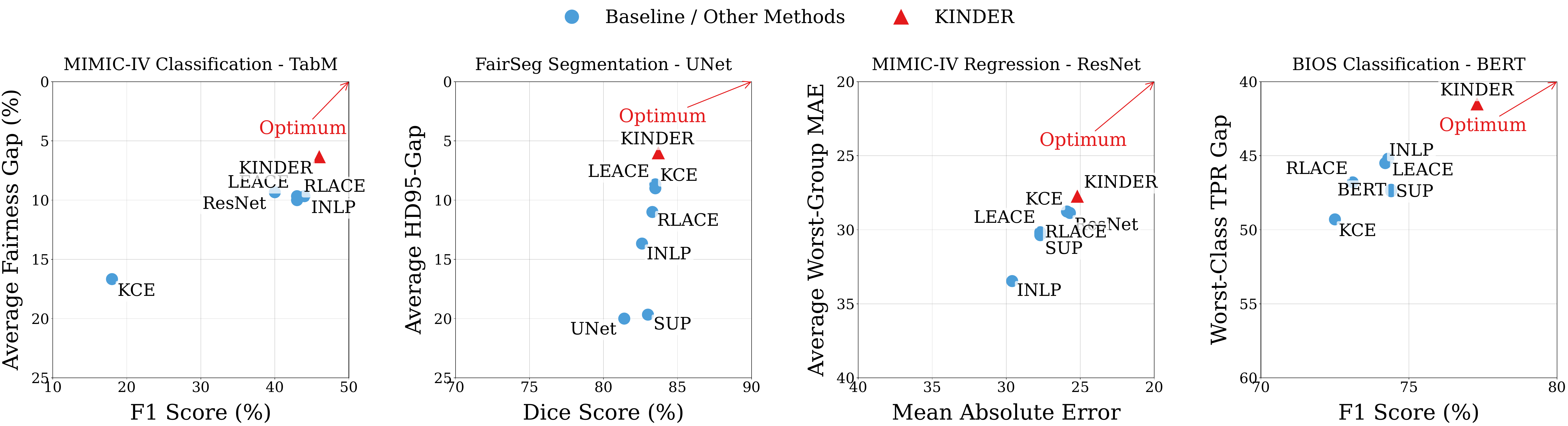}
    \vspace{-5mm}
    \caption{
    \textbf{Performance-fairness tradeoff plots} across classification, segmentation \& regression.
    %, and BIOS classification. 
    \textcolor{red}{\textbf{Red}} triangles denote \methodname; 
    \textcolor{MidnightBlue}{\textbf{blue}} circles denote baselines and competing post-hoc methods.
    }
    \label{fig:combined_fairness_plots}
    \vspace{-5mm}
\end{figure*}

\vspace{-3mm}
\section{Experiment settings}
\vspace{-2mm}

%\paragraph{Datasets}

\textbf{Datasets used:}
\textit{Supervised classification:}
Bias in Bios \cite{de2019bias} and
Moji \cite{zhou2018mojitalk} (for post-hoc evaluation),
MIMIC-IV \cite{johnson2020mimic},
\textit{Supervised regression:}
Communities and Crime \cite{mcclendon2015using},
MIMIC-IV \cite{johnson2020mimic},
\textit{Supervised segmentation:}
FairSeg \cite{tianfairseg},
\textit{SSL classification:}
Harvard-FairVLMed \cite{luo2024fairclip} (multimodal).
%
%%%
CelebA \cite{liu2015deep} (unimodal).
% ACSIncome \cite{ding2021retiring},

%
%Please refer to Appendix \ref{sect:appendix_experimental_setting} for full  dataset details.
%
%\hl{
%}
%
% \textbf{Fairness metrics:} 
% For segmentation, we report 
% worst-group Dice for overlap fairness \citep{perez2025exploring}, and HD95 gap for boundary fairness \citep{vcevora2024quantifying}. 
% %
% For text classification, we report worst-group macro-F1 \citep{chalkidis2022fairlex}, worst-class TPR gap \citep{romanov2019s,hemamou2022delivering}, and RMS class TPR gap \citep{ravfogel2020null,ravfogel2022linear}. 
% %
% For other classification tasks, we report the aggregated fairness score following 
% \citep{cheong2025fairwell}, calculated as the average of statistical parity, equal accuracy, equalized odds, and equal opportunity. 
% For regression, we report MAE gap for group error disparity \citep{han2022systematic}.
% %
% We provide all the further training, datasets, implementation, and fairness measure details within the Appendix (Appendix \ref{sect:appendix_experimental_setting}). 
% We select hyperparameters by validation task utility (AUC for classification, MAE for regression, and Dice for segmentation), rather than fairness or erasure scores. Our main research target is exploring how much protected-attribute dependence can be removed while preserving high downstream performance.

\textbf{Fairness metrics:}
For segmentation, we report worst-group Dice for overlap fairness \citep{perez2025exploring} and HD95 gap for boundary fairness \citep{vcevora2024quantifying}. For text classification, we report worst-group macro-F1 \citep{chalkidis2022fairlex}, worst-class TPR gap \citep{romanov2019s,hemamou2022delivering}, and RMS class TPR gap \citep{ravfogel2020null,ravfogel2022linear}. For other classification tasks, we use the aggregated fairness score from \citep{cheong2025fairwell,fairssl}, averaging statistical parity, equal accuracy, equalized odds, and equal opportunity. For regression, we report MAE gap for group error disparity \citep{han2022systematic}. Further training, dataset, implementation, and %fairness-measure 
details are provided in Appendix~\ref{sect:appendix_experimental_setting}. 
Hyperparameters are selected by validation utility. For \methodname, we fix the Gaussian RBF kernel and moment-based prototype family, while selecting the prototype order by the validation criterion (Appendix~\ref{sect:appendix_implementation_details}). Our goal is to evaluate how much protected-attribute dependence can be removed while preserving performance.

\vspace{-2mm}
\section{Results}
\label{sect:results}
\vspace{-2mm}

\begin{table}[t]
%\begin{wraptable}{r}{0.72\textwidth}
\vspace{-19.5pt}
\centering
%\scriptsize
\footnotesize
\caption{\textbf{BIOS \& MOJI classification results.} 
Abbreviations:
WgF1: worst-group macro-F1. WCTG: worst-class TPR gap. RCTG: RMS class TPR gap. Arrows indicate whether higher or lower values are better.
\label{tab:bios_moji_results} 
First row is the baseline model. Rows prefixed with ``+'' apply the corresponding method on top of the baseline. %backbone. 
Best values %are 
in \textbf{bold}, second-best values %are 
\underline{underlined} for all tables.
}
\vspace{-1mm}
\setlength{\tabcolsep}{2.5pt}
\setlength{\tabcolsep}{1.8pt}
\begin{tabular}{@{}lcccccccccc@{}}
\toprule
\multirow{2}{*}{\textbf{Method}}
& \multicolumn{5}{c}{\textbf{BIOS}}
& \multicolumn{5}{c}{\textbf{MOJI}} \\
\cmidrule(lr){2-6}
\cmidrule(lr){7-11}
& \textbf{Acc $\uparrow$}
& \textbf{F1 $\uparrow$}
& \textbf{WgF1 $\uparrow$}
& \textbf{WCTG $\downarrow$}
& \textbf{RCTG $\downarrow$}
& \textbf{Acc $\uparrow$}
& \textbf{F1 $\uparrow$}
& \textbf{WgF1 $\uparrow$}
& \textbf{WCTG $\downarrow$}
& \textbf{RCTG $\downarrow$} \\
\midrule
BERT
& \underline{80.9} & \underline{74.4} & 70.7 & 47.3 & 16.2
& \textbf{80.1} & 74.7 & \underline{62.1} & 25.6 & 17.5 \\

+ INLP
& 80.7 & 74.2 & 70.3 & 45.5 & 14.9
& 79.8 & 74.7 & 61.5 & 25.3 & 17.9 \\

+ RLACE
& 79.8 & 73.1 & 69.1 & 46.8 & \underline{14.8}
& 79.1 & 73.7 & 61.0 & 23.9 & 16.9 \\

+ LEACE
& 80.7 & 74.3 & 70.5 & \underline{45.2} & \textbf{14.5}
& \underline{79.9} & 74.8 & 61.1 & 26.3 & 18.6 \\

+ KCE
& 79.8 & 72.5 & 68.6 & 49.3 & 16.5
& 79.2 & 73.9 & 61.4 & 23.6 & \underline{16.7} \\

+ SUP
& 80.8 & \underline{74.4} & \underline{71.0} & 47.4 & 15.9
& 79.8 & 74.6 & 61.8 & \underline{23.8} & 16.8 \\

\cmidrule(lr){1-11}
+ \methodname
& \textbf{81.9} & \textbf{77.3} & \textbf{73.9} & \textbf{41.5} & 15.2
& 79.2 & \textbf{77.5} & \textbf{65.3} & \textbf{16.7} & \textbf{14.1} \\
\bottomrule
\end{tabular}
\end{table}
%\vspace{-0.4cm}
%\end{wraptable}

Across Tables~\ref{tab:bios_moji_results}--\ref{tab:fairvlmed_ssl_combined}, we evaluate whether \methodname remains effective under dataset imbalance (Appendix~\ref{sect:appendix_datasets}, Table~\ref{tab:dataset_split_imbalance}) while preserving utility and improving fairness across single-/multiple-attribute, intersectional, supervised, and self-supervised settings.%
%\subsection{Classification results.}
%Across Tables~\ref{tab:bios_moji_results}--\ref{tab:fairvlmed_ssl_combined}, we evaluate whether \methodname preserves utility while improving fairness across single-attribute, multiple-attribute, intersectional, supervised, and self-supervised settings.
\vspace{-3mm}
\paragraph{Single attribute classification results.}
%
%We first evaluate classification settings where the target label is predictive, but the learned representation may also encode sensitive group structure. 
%
%These experiments compare \methodname with projection-based and adversarial erasure baselines that define removal through linear probes, minimax games, closed-form linear erasure, or a chosen kernelized adversary \cite{ravfogel2020null,ravfogel2022linear,belrose2023leace,ravfogel2022adversarial}. 
%
%The main question is whether estimating the nuisance subspace from group prototypes gives a better utility--fairness trade-off than erasure procedures tied to a specific recovery family.
%
%
% \methodname (LR head) & \textbf{84.0} & 76.6 & 45.4 & 14.9  & \underline{83.5} & \underline{66.8} & \underline{23.0} & \underline{16.5} \\

%
%
%
On 
\underline{BIOS and MOJI} (Table~\ref{tab:bios_moji_results}), which evaluate \methodname in post-hoc settings, \methodname gives the strongest fairness-sensitive classification results. On BIOS, it improves accuracy and worst-group F1 while reducing the worst-class TPR gap. On MOJI, the accuracy is slightly lower than the BERT baseline, but F1 worst-group, F1, and class-level disparity metrics improve by a larger margin. 
%
%These two cases show the same general behavior under different utility levels: 
%
%\methodname does not only suppress a probe-specific notion of recoverability, but produces cleaned representations that are more stable across protected groups.

\begin{table}[t]
\centering
\caption{\textbf{FairSeg segmentation} results on UNet. Dice and HD95 measure performance, Dice measures overlap and HD95 measures boundary error. D-WG and H-Gap measure fairness, where D-WG denotes worst-group Dice and H-Gap denotes HD95 gap. %Arrows indicate the preferred direction. 
%Best values are in \textbf{bold} and second-best values are \underline{underlined}.
\label{tab:fairseg_transunet}}
\vspace{-2mm}
\small
\scalebox{1.0}{
\resizebox{\textwidth}{!}{
\begin{tabular}{lcc|cc|cc|cc|cc}
\toprule
\multirow{2}{*}{\textbf{Method}}
& \multicolumn{2}{c|}{\textbf{Perf.}}
& \multicolumn{2}{c|}{\textbf{Race}}
& \multicolumn{2}{c|}{\textbf{Gender}}
& \multicolumn{2}{c|}{\textbf{Ethnicity}}
& \multicolumn{2}{c}{\textbf{Inter.}} \\
\cmidrule(lr){2-3}
\cmidrule(lr){4-5}
\cmidrule(lr){6-7}
\cmidrule(lr){8-9}
\cmidrule(lr){10-11}
& \textbf{Dice $\uparrow$}
& \textbf{HD95 $\downarrow$}
& \textbf{D-WG $\uparrow$}
& \textbf{H-Gap $\downarrow$}
& \textbf{D-WG $\uparrow$}
& \textbf{H-Gap $\downarrow$}
& \textbf{D-WG $\uparrow$}
& \textbf{H-Gap $\downarrow$}
& \textbf{D-WG $\uparrow$}
& \textbf{H-Gap $\downarrow$} \\
\midrule
UNet & 81.4 & 6.8 & 77.4 & 4.1 & 81.3 & 0.7 & 81.3 & 1.2 & 67.5 & 30.7 \\
+ INLP    & 82.6 & 5.5 & 78.7 & 3.0 & 82.6 & 1.0 & 82.6 & \textbf{0.1} & 70.5 & 20.0 \\
+ RLACE   & 83.3 & 5.1 & 80.0 & 2.8 & 83.2 & \underline{0.3} & 83.2 & \underline{0.2} & 73.1 & 17.0 \\
+ LEACE   & 83.5 & 4.7 & 80.3 & \textbf{1.2} & 83.5 & 0.6 & 83.4 & 0.8 & \underline{76.6} & \underline{6.0} \\
+ SUP     & 83.0 & 5.5 & 79.1 & 3.7 & 83.0 & 0.8 & 82.9 & 1.4 & 68.7 & 30.8 \\
+ KCE     & 83.5 & \underline{4.5} & 80.2 & 1.9 & 83.3 & \underline{0.3} & 83.4 & 0.5 & 75.7 & 6.9 \\
\cmidrule(lr){1-11}
+ \methodname  & \textbf{83.7} & \textbf{4.1} & \textbf{80.4} & \underline{1.6} & \textbf{83.7} & \textbf{0.1} & \textbf{83.7} & \textbf{0.1} & \textbf{78.2} & \textbf{5.4} \\
\bottomrule
\end{tabular}
}
}
\vspace{-0.5cm}
\end{table}

\begin{table*}[t]
\centering
\setlength{\belowcaptionskip}{4pt}
\caption{\textbf{MIMIC-IV classification} results using TabM. Acc and F1 are performance metrics and the remaining columns are fairness metrics. Arrows indicate the preferred direction. \methodname (NH) uses the native end-to-end task head, while \methodname replaces this head with a linear probe as an ablation comparison with other post-hoc erasure baselines. Post-hoc ablations are reported in Appendix~\ref{app:mimic_seq_ablation}.}
\label{tab:mimiciv_group_intersectional}
\footnotesize
\scalebox{0.95}{
\resizebox{\textwidth}{!}{%
\begin{tabular}{lccccccc}
\toprule
\textbf{Method}
& \textbf{Acc $\uparrow$}
& \textbf{F1 $\uparrow$}
& \textbf{Gender Gap $\downarrow$}
& \textbf{Race Gap $\downarrow$}
& \textbf{Age Gap $\downarrow$}
& \textbf{Inter. Gap $\downarrow$}
& \textbf{WG Inter.  $\uparrow$} \\
\midrule
TabM      & \underline{91.9} & 39.9 & 2.1 & 23.2 & \textbf{2.5} & \underline{46.6} & \underline{65.8} \\
+ INLP    & 79.5 & 43.6 & 1.6 & 22.1 & 5.2 & 53.3 & 58.3 \\
+ RLACE   & 78.5 & 43.0 & 2.4 & \underline{20.7} & 7.1 & 53.1 & 58.3 \\
+ LEACE   & 78.3 & 42.6 & 2.3 & 21.4 & 6.2 & 54.4 & 58.3 \\
+ KCE     & 80.1 & 18.0 & \textbf{0.4} & 46.2 & 4.1 & 73.6 & 35.7 \\
\hline
+ \methodname & 80.6 & \underline{44.7} & 1.2 & 22.5 & 7.1 & \underline{46.6} & \textbf{68.7} \\
+ \methodname (NH)     & \textbf{92.0} & \textbf{45.5} & \underline{0.7} & \textbf{15.4} & \underline{3.3} & \textbf{44.8} & \textbf{68.7} \\
\bottomrule
\end{tabular}
}
}
\vspace{-2mm}
\end{table*}

\vspace{-3mm}
\paragraph{Intersectional fairness results.}
A similar pattern appears on 
\underline{MIMIC-IV} (Table~\ref{tab:mimiciv_group_intersectional}), 
where we test the same claim in a setting with multiple protected attributes.
%and intersectional groups. 
%
The joint \methodname variant gives the best overall balance across the table. 
It preserves the baseline accuracy, improves F1, reduces the race gap, and improves worst-group accuracy. 
%
%Several post-hoc baselines improve individual gap metrics, but they also reduce accuracy, F1, or worst-group accuracy more sharply. 
%
%This setting is important because clinical fairness is rarely a single-attribute problem. 
%
%A method can reduce one demographic gap while still leaving large disparities for another attribute or for an intersectional subgroup.  
%
Further multi-attribute ablations %on the results 
of sequential attributes removal over the baseline methods are provided in Appendix~\ref{app:mimic_seq_ablation}.
%
%The MIMIC-IV results therefore test whether the same erasure operator remains stable when several protected group structures are present at the same time.

%\begin{table}[h]
\begin{wraptable}{r}{0.70\textwidth}
\vspace{-6.1mm}
\centering
\caption{I-JEPA results on CelebA. Acc. and WG-Inter. measure utility and worst-group intersectional performance, where higher is better; gap metrics measure protected-attribute disparity, where lower is better.}
\label{tab:celeba_ijepa_results}
\small
\setlength{\tabcolsep}{3.2pt}
\resizebox{0.66\textwidth}{!}{%
\begin{tabular}{lccccc}
\toprule
Method & Acc. $\uparrow$ & Gender Gap $\downarrow$ & Age Gap $\downarrow$ & Inter. Gap $\downarrow$ & WG-Inter. $\uparrow$ \\
\midrule
I-JEPA      & 91.1 & 6.6 & 2.7 & 11.5 & 87.7 \\
+ INLP      & 85.7 & \textbf{3.4} & 8.5 & 13.7 & 82.7 \\
+ LEACE     & 90.0 & 6.3 & 5.8 & 12.0 & 86.3 \\
+ RLACE     & 91.8 & 6.1 & 2.6 & \textbf{10.8} & 88.5 \\ 
+ KCE       & 91.8 & 7.0 & \underline{2.5} & 11.4 & 88.3 \\
+ SUP       & \underline{91.9} & \underline{6.0} & 2.7 & \underline{11.0} & 88.5 \\
+ \methodname & \textbf{92.2} & 6.3 & \textbf{2.3} & 11.4 & \textbf{89.4} \\
\bottomrule
\end{tabular}
}
\end{wraptable}
%\end{table}

%\textcolor{teal}{
In the MIMIC-IV classification results we observe the potential issues that might arise from head mismatch in post-hoc projection methods. When the native \methodname head is replaced with linear probing for comparability with prior erasure baselines, performance drops sharply even though fairness remains competitive. The gap between \methodname (NH) and \methodname suggests that replacing the original task head can impose a substantial utility bottleneck on some datasets. Similar observations were made in our sequential attribute removal ablation in Appendix~\ref{app:mimic_seq_ablation}. This is the setting in which an end-to-end compatible erasure operator is practically advantageous. 
%}

\paragraph{Additional three-seed results.}
Additional three-seed evaluations on BIOS, MIMIC-IV classification and regression, and FairSeg are reported in Appendix~\ref{app:rebuttal_three_seed}. Among projection-based baselines, we provide additional analysis in Appendix~\ref{app:rebuttal_first_order_sal} on SAL~\cite{shao2023gold}, as it uses covariance-based spectral erasure with a kernelized extension, which motivated a separate comparison with first-order \methodname. We observe that \methodname's advantages remain against all these baselines.

\vspace{-4mm}
\paragraph{Matched end-to-end results.}
On MIMIC-IV classification, we further compare \methodname (NH) with FR-Train, FairMixup, GroupDRO, and FairBatch using TabM as the backbone (Appendix~\ref{app:rebuttal_matched_end_to_end}). \methodname achieves the highest F1 (46.17), compared with 44.66 for FairMixup and 41.72 for TabM. The results show that the advantage of \methodname remains under completely matched end-to-end fairness-specific baselines.

%%%%%%%%%%%%%%%%%%%%%%%%%%%%%%%%%%%%%%%%%%

%\subsection{Segmentation and regression results.}

% \begin{table}[t]
% \centering
% \caption{FairSeg: Segmentation task \label{tab:fairseg_segmentation}}
% \small
% \setlength{\tabcolsep}{5pt}
% \renewcommand{\arraystretch}{1.15}
% \scalebox{0.85}{
% \begin{tabular}{|c|c|c|c|c|c|c|c|c|c|}
% \hline
% \multicolumn{1}{|c|}{Backbone} & \multicolumn{1}{c|}{Type} & \multicolumn{2}{c|}{Performance} & \multicolumn{2}{c|}{Race} & \multicolumn{2}{c|}{Gender} & \multicolumn{2}{c|}{Ethnicity} \\
% \cline{3-10}
% & & Dice$\uparrow$ & HD95$\downarrow$ & ES-Dice$\uparrow$ & HD95 gap$\downarrow$ & ES-Dice$\uparrow$ & HD95 gap$\downarrow$ & ES-Dice$\uparrow$ & HD95 gap$\downarrow$ \\
% \hline
% \raisebox{-1.5ex}[0pt][0pt]{UNet} & Baseline & 0.84 & 4.12 & 0.78 & 1.60 & 0.84 & 0.26 & 0.82 & 0.77 \\
% \cline{2-10}
% & \methodname & 0.84 & \textbf{3.97} & 0.78 & \textbf{1.13} & 0.84 & \textbf{0.10} & \textbf{0.83} & \textbf{0.53} \\
% \hline
% \raisebox{-1.5ex}[0pt][0pt]{TransUNet} & Baseline & 0.84 & 4.88 & 0.78 & 1.97 & 0.83 & 0.22 & 0.82 & 1.17 \\
% \cline{2-10}
% & \methodname & 0.84 & \textbf{4.25} & \textbf{0.79} & \textbf{1.50} & \textbf{0.84} & \textbf{0.04} & \textbf{0.83} & \textbf{0.79} \\
% \hline
% \end{tabular}
% }
% \end{table}

%%%%%%%%%%%%%
%

\begin{table}[t]
\centering
\caption{\textbf{MIMIC-IV regression results on ResNet and MLP-PLR.} MAE and $R^2$ measure performance; worst-group MAE columns measure fairness. Arrows indicate the preferred direction. Best values are in \textbf{bold} and second-best values are \underline{underlined}.}
\vspace{-1mm}
\label{tab:mimic_combined_summary_wg}
\footnotesize
\scalebox{0.95}{
\resizebox{\columnwidth}{!}{%
\begin{tabular}{lcccccc}
\toprule
\textbf{Method}
& \textbf{MAE $\downarrow$}
& \textbf{$R^2$ $\uparrow$}
& \textbf{Race WG-MAE $\downarrow$}
& \textbf{Gender WG-MAE $\downarrow$}
& \textbf{Age WG-MAE $\downarrow$}
& \textbf{Inter. WG-MAE $\downarrow$} \\
\midrule
ResNet & 2.59 & \textbf{0.15} & 3.24 & 2.61 & \underline{2.78} & \underline{5.55} \\
+ INLP   & 2.96 & 0.00 & 3.75 & 3.01 & 3.28 & 10.88 \\
+ LEACE  & 2.77 & 0.11 & 3.25 & 2.79 & 3.01 & 5.91 \\
+ RLACE  & 2.77 & 0.11 & 3.27 & 2.79 & 3.02 & 5.93 \\
+ KCE    & \underline{2.57} & \textbf{0.15} & \underline{3.23} & \underline{2.59} & 2.84 & \textbf{5.10} \\
+ SUP    & 2.77 & 0.11 & 3.29 & 2.79 & 3.03 & 5.90 \\
\midrule
+ \methodname & \textbf{2.52} & 0.13 & \textbf{3.07} & \textbf{2.53} & \textbf{2.72} & 7.01 \\
\midrule
MLP-PLR & \textbf{2.59} & 0.11 & \underline{3.35} & \underline{2.62} & \underline{2.84} & 5.65 \\
+ INLP    & 2.67 & 0.11 & 3.45 & 2.72 & \textbf{2.83} & 5.87 \\
+ LEACE   & 2.69 & 0.09 & 3.40 & 2.73 & \textbf{2.83} & 5.25 \\
+ RLACE   & 2.68 & 0.09 & 3.43 & 2.71 & 2.89 & \textbf{5.05} \\
+ KCE     & 2.67 & 0.10 & 3.50 & 2.70 & 2.89 & 5.29 \\
+ SUP     & 2.68 & 0.09 & 3.50 & 2.72 & 2.88 & \underline{5.22} \\
\midrule
+ \methodname & \textbf{2.59} & \textbf{0.12} & \textbf{3.30} & \textbf{2.61} & 2.85 & 6.83 \\
\bottomrule
\end{tabular}
}%
}
\vspace{-4mm}
\end{table}

\vspace{-2mm}
\begin{table}[t]
\centering
\caption{\textbf{FairVLMed multimodal self-supervised learning results} on CLIP and FactorCL. Accuracy measures performance; gap and worst group intersectional accuracy measures fairness. Arrows indicate the preferred direction. 
%Best values are in \textbf{bold} and second-best values are \underline{underlined}.
}
\vspace{-1mm}
\small
% \begin{tabular}{lrrrrrrr} % Age Gap commented out
\scalebox{0.90}{
\begin{tabular}{lcccccc} 
\toprule
\textbf{Method} 
& \textbf{Acc $\uparrow$}
& \textbf{Race Gap $\downarrow$}
& \textbf{Gender Gap $\downarrow$}
% & \textbf{Age Gap $\downarrow$}
& \textbf{Ethnicity Gap $\downarrow$}
& \textbf{Inter. Gap $\downarrow$}
& \textbf{WG-Inter. $\uparrow$} \\
\midrule
FactorCL  & 60.4 & 12.8 & 4.3 % & 24.1
& 15.6 & 95.8 & 0.0 \\
+ INLP         & 60.2 & 7.9 & 5.6 % & \underline{6.3}
& \underline{10.3} & \textbf{76.0} & \underline{25.0} \\
+ RLACE        & 60.0 & 13.7 & 4.2 % & 22.9
& 16.9 & 95.9 & 0.0 \\
+ LEACE        & 57.3 & \textbf{4.3} & 3.7 % & \textbf{3.1}
& \textbf{7.3} & 84.8 & 0.0 \\
+ KCE          & 58.8 & 9.1 & 5.8 % & 16.1
& 15.7 & 93.5 & 0.0 \\
+ SUP          & \underline{60.6} & 12.3 & \textbf{2.3} % & 21.7
& 17.7 & 95.8 & 0.0 \\
% \cmidrule(lr){1-8} % Age Gap commented out
\cmidrule(lr){1-7}
+ \methodname       & \textbf{61.2} & \underline{6.9} & \underline{3.0} % & 11.2
& 12.4 & \underline{83.7} & \textbf{41.5} \\
\midrule
CLIP      & \textbf{73.6} & \underline{6.2} & 4.0 % & 18.5
& 14.2 & 80.7 & 41.9 \\
+ INLP         & 70.5 & 12.0 & 5.4 % & \underline{5.3}
& 13.5 & \textbf{74.6} & \textbf{50.0} \\
+ RLACE        & 73.0 & 7.4 & \underline{2.8} % & 18.3
& 16.1 & 80.7 & 41.9 \\
+ LEACE        & 69.3 & 6.7 & 6.2 % & \textbf{4.0}
& 18.0 & 87.1 & 0.0 \\
+ KCE          & \underline{73.4} & 6.3 & 4.1 % & 25.4
& 18.9 & 83.6 & 44.5 \\
+ SUP          & \underline{73.4} & 6.3 & 3.1 % & 19.1
& \underline{12.8} & 80.2 & 44.9 \\
% \cmidrule(lr){1-8} % Age Gap commented out
\cmidrule(lr){1-7}
+ \methodname       & 71.6 & \textbf{5.7} & \textbf{1.6} % & 16.2
& \textbf{5.9} & \underline{79.1} & \textbf{50.0} \\
\bottomrule
\end{tabular}
}
\vspace{-5mm}
\label{tab:fairvlmed_ssl_combined}
\end{table}

\vspace{-3mm}
\paragraph{Segmentation results.}
On FairSeg (Table~\ref{tab:fairseg_transunet}), 
%\methodname extends the same pattern to dense prediction. T
the setting is different from classification as fairness improvements cannot be attributed only to threshold shifts or changes in class scores and the method must preserve key spatial information as well. 
\methodname improves the overall Dice and HD95 scores relative to the UNet baseline, while improving worst-group Dice and reducing the largest boundary disparities. The strongest gains appear on the intersectional metrics, where the baseline has the largest gap. This supports that \methodname is not tied to any particular classifier head or task loss.

\vspace{-3mm}
\paragraph{Regression results.}
%
%The regression results give a less direct but still consistent picture. 
On MIMIC-IV regression (Table~\ref{tab:mimic_combined_summary_wg}), \methodname improves ResNet MAE and marginal worst-group errors, with intersectional worst-group MAE as the main exception. Similar observations appear for MLP-PLR in the same table. On Communities and Crime (Appendix Table~\ref{tab:communities_crime_mlp_plr_race_summary}), \methodname gives the best utility and worst-group error, while remaining close to the strongest race-gap result. This is important as some post-hoc baselines, especially INLP and LEACE, reduce the representation aggressively and lose a large amount of predictive information.

\vspace{-1mm}
\paragraph{Multimodal SSL results.}
%
%Table~\ref{tab:fairvlmed_ssl_combined} evaluates whether the same erasure operator remains useful in multimodal self-supervised learning. This is a key setting for \methodname, because the block is applied during representation learning rather than only as a post-hoc transform on fixed supervised features. 
%
Looking at FairVLMed (Table~\ref{tab:fairvlmed_ssl_combined}), on FactorCL, \methodname improves accuracy while reducing all reported gaps and gives the strongest worst-intersectional balanced accuracy. On CLIP, \methodname gives the best race, gender, and ethnicity gaps, and matches the best worst-intersectional balanced accuracy. CelebA (Table~\ref{tab:celeba_ijepa_results}) shows the same unimodal SSL pattern: \methodname gives the best accuracy, age gap, and worst-intersectional performance, while keeping gender and intersectional gaps close to the baseline.
%
%but the accuracy is lower than the unmodified CLIP baseline. 
%
%This suggests that the effect of erasure depends on how strongly the pretrained representation entangles task-relevant and group-associated structure. 
%
%These observations support the paper's main claim. We observe that \methodname remains compatible with multimodal SSL and can reduce several protected-attribute gaps, but the utility cost also depends on the representation geometry.
\vspace{-2mm}
\section{Discussion and Conclusion}
\label{sect:discussion}

%%%%%%%%%%%%%%%%%%%%
\vspace{-2mm}
\paragraph{RQ 1: Representation-level estimation:}
We show that in \methodname,
a projection-based operator is able to 
estimate erasable group-conditioned structure directly at the representation level by estimating the removable structure from the explicit protected-group prototypes in a lifted prototype space.
The sequential SVD then allows higher-order prototypes to contribute residual nuisance directions beyond those identified at lower orders, without relying on a selected probe or adversary to define the erasure target. Our prototype-order ablation (App.~\ref{app:prototype_order_ablation}) further shows that higher orders reduce measured feature dependence and higher-order group discrepancies, while increasing the projection shift.

%%%%%%%%%%%%%%%%%%%%
\vspace{-2mm}
\paragraph{RQ 2: Generalizable concept erasure:}
%
%We introduced \methodname, a projection-based concept-erasure operator that removes group-conditioned nuisance structure directly at the level of intermediate representations. 
%
As a result, \methodname is able to achieve \textit{generalizable concept erasure}
%
%Our results indicate that it is possible to achieve generalizable concept erasure via 
by moving the target of removal from a restricted decoder family to the cleaned representation itself.
\textit{Empirically}, \methodname is shown to work across a variety of training paradigms (i.e. during training and post training) and architectures (BERT, IJEPA, ResNet, UNet, CLIP, etc.).
\textit{Theoretically}, our analysis shows that 
%under a shared-base structural model with group-specific perturbations, the projection (ref Sec. \ref{sect:method_projection})
\methodname
removes the nuisance directions responsible for protected-group variation and yields independence from the protected attribute at the level identified by the prototype space. 
Further, under stronger prototype-exactness conditions, our prototype-level guarantee recovers probabilistic independence and thus, it separates the main target of erasure from a particular decoder family while preserving the explicit structure of projection-based concept erasure. Under such conditions, independence is also preserved under any measurable downstream transformation of the cleaned representation.

\begin{comment}

This is evident across our results which works across a variety of training paradigms (i.e. during training and post training) and 
architectures (BERT, IJEPA, LEJEPA, ResNet, UNet etc.).
%
This is further supported by our theoretical analysis proving that 
non-recoverability for all downstream (measurable) decoders for the selected
space follows automatically from \methodname.
%
Our theoretical analysis shows that under a shared-base structural model with group-specific perturbations, the projection (ref Sec. \ref{sect:method_projection})
removes the nuisance directions responsible for protected-group variation and yields independence from the protected attribute at the level identified by the prototype space. 
%
Further, under stronger kernel-exactness conditions, our prototype-level guarantee recovers full probabilistic independence, thus separating the main target of erasure from a particular decoder family while preserving the explicit structure of projection-based concept erasure which leads to \textit{highly generalizable concept erasure}.
%
%Instead of defining erasure through the failure of a selected probe, adversary, or recovery family, 
%
%\methodname estimates the removable structure from explicit protected-group prototypes in a lifted prototype space and applies an orthogonal projection to obtain cleaned representations. 
%

\end{comment}

%This allows \methodname to be compatable with a wide range of 

%%%%%%%%%%%%%%%%%%%%
\vspace{-2mm}
\paragraph{RQ 3: All round utility across a variety of training paradigms, tasks and settings:}
\methodname improves the empirical fairness--utility tradeoff under our utility-first selection protocol \cite{zhao2022inherent,chowdhury2025fundamental}.
During end-to-end training, the encoder can adapt to the projection and retain task-relevant signals through the remaining representation directions, rather than relying on the removed nuisance structure.
\methodname achieves strong performance across 
%diverse settings, including 
supervised (e.g., MIMIC-IV dataset, Bios \& Moji), self-supervised (e.g., FairVLMed, CelebA), unimodal, and multimodal settings,
%without modifying 
%changing the downstream objective, thus 
and consistently improves both fairness and performance 
(see Fig. \ref{fig:combined_fairness_plots})
for a variety of tasks
even under \textit{multiple} and \textit{intersectional} protected attributes. 
The computational-cost analysis further shows that \methodname maintains competitive runtime and memory requirements compared with other concept erasure methods (Appendix Table~\ref{tab:rebuttal_compute_cost}).
Our findings support %prototype-space erasure 
\methodname
as an all-round
practical and theoretically grounded approach to %decoder-agnostic 
\textit{generalizable task-agnostic fair representation learning}.

\vspace{-2mm}
\paragraph{Limitations:}
Our experiments are restricted to datasets that contain sensitive attributes which is the common setting in fairness analysis. We only focus on \methodname's moment order functionals and not other forms of functionals. Future work can extend the analysis on other datasets and new variations of prototype functionals.
We provide an extended discussion of the limitations in Appendix \ref{sect:appendix_limitation}.

%utility-fairness trade-offs (see Fig \ref{fig:combined_fairness_plots}). across various classification, segmentation, and regression tasks in both supervised and self-supervised learning settings. 
%
%The results show that the same erasure operator can improve group robustness across various modalities,
%
%They also show the practical value of using \methodname as an in-model erasure block. When the cleaned representation remains paired with a compatible task head, 
%
%We observe that utility is often better preserved than in post-hoc linear probing settings (see Table \ref{tab:mimiciv_group_intersectional})
%
%
%
%%%%%%%%%%%%
%
\begin{comment}
%
This is evident across our results which show that
the proposed
simple representation-cleaning block %that 
can be inserted into 
standard supervised (e.g. MIMIC-IV, Bios \& Moji), 
self-supervised (e.g. FairVLMed), 
unimodal (e.g. MIMIC-IV, Bios \& Moji), and 
multimodal (e.g. FairVLMed) 
settings without changing the downstream objective, thus resulting in improved fairness and performance \ref{fig:combined_fairness_plots}. 
%
\end{comment}
%
%%%%%%%%%%%%

\vspace{-2mm}

\begin{ack}
\vspace{-1mm}

Abtin Mogharabin is supported by the Konrad Zuse School of Excellence in Learning and Intelligent Systems (\href{https://eliza.school/}{ELIZA}) through the DAAD programme Konrad Zuse Schools of Excellence in Artificial Intelligence, sponsored by the Federal Ministry of Education and Research.
Jiaee Cheong is supported by the DAAD Fellowship and the Harvard BKC Fellowship.
We also gratefully acknowledge the computational resources kindly provided by METU Image Lab and METU-ROMER, Center for Robotics and Artificial Intelligence.
\end{ack}

%\clearpage
\bibliographystyle{plain} % Or the specific NeurIPS style
\bibliography{paper,neurips26}      % Do NOT use .bib extension here

\newpage

\appendix

%\section*{Appendices} 

% Tag the start of the appendix
\etocdepthtag.toc{appendix}

% Tell etoc to hide the 'main' tag and only show the 'appendix' tag
\etocsettagdepth{main}{none}
\etocsettagdepth{appendix}{subsection} % Change to 'section' if you don't want subsections

% Set a custom title for the local TOC
\etocsettocstyle{\subsection*{Table of Contents: Appendix}}{}

\tableofcontents

\renewcommand\thefigure{\thesection.\arabic{figure}}   
\renewcommand\thetable{\thesection.\arabic{table}}
\renewcommand\theequation{\thesection.\arabic{equation}}

\section{More Related Work}
\label{app:additional_related_work}

\paragraph{Projection-based concept-erasure methods.}
INLP relies on iterative linear probes and a joint-nullspace projection for concept erasure \cite{ravfogel2020null}. 
RLACE formulates concept erasure as a linear minimax game where the concept is removed from the learned subspace \cite{ravfogel2022linear}. 
KCE kernelizes the linear minimax game and provides erasure against a selected kernelized adversary \cite{ravfogel2022adversarial}. 
LEACE is a closed-form method that prevents all linear classifiers from detecting a concept while changing the embedding as little as possible \cite{belrose2023leace}. 
SUP obtains debiased representations by projecting them onto the subspace orthogonal to a sufficient span \cite{shi2024debiasing}. 
QLEACE extends LEACE by removing quadratically available information about a concept from a representation \cite{quirke2025slowing}. 
Obliviator targets nonlinear guardedness by minimizing statistical dependence between learned PLM representations and unwanted attributes \cite{akbari2026obliviator}. 
These methods motivate Table~\ref{tab:comparative_summary}, where the main distinction is whether erasure is defined through a restricted decoder/adversary family or directly at the cleaned-representation level.

%%%%%%%%%%%%%%%%%%%%%%%%%%%%%%%%%%%%%%%%%%%%%%%%%%%%%%%%%%%%%%%%%%%%%%%%%%%

%%%%%%%%%%%%%%%%%%%%%%%%%%%%%%%%%%%%%%%%%%%%%%%%%%%%%%%%%%%%%%%%%%%%%%%%%%%

\section{Further \methodname Details}
\label{sect:appendix_further_kinder_details}

\subsection{\methodname Algorithm}
\label{sect:appendix_KINDER_algorithm}

We present the full algorithm in Algorithm \ref{alg:methodname_prototypes}.

\begin{algorithm}[t]
\caption{\methodname for a batch: first-order prototype subspace projection.
}
\label{alg:methodname_prototypes}
\begin{algorithmic}[1]
\Require Minibatch representations $\mathbf Z\in\mathbb R^{B\times d}$ and protected-group labels $\mathbf s=(s^{(1)},\ldots,s^{(B)})$, with $s^{(i)}\in\{1,\ldots,G\}$
\Require RFF feature map $\Phi_\mathrm{rff}$ defined in Eq.~\eqref{eq:rff}
\Require Removal rank $d_N\le G-1$, EMA rate $\rho\in(0,1]$
\Require Prototype-subspace refresh interval $T_{\mathrm{refresh}}\in\mathbb N$ (measured in optimization steps) and current optimization step $t_{\mathrm{step}}\in\mathbb N$
\Require Empirical group frequencies $(\hat\pi_1,\ldots,\hat\pi_G)$
\Require Stored mean prototypes $\{\mu_s\in\mathbb R^{M}\}_{s=1}^{G}$ and prototype basis $\mathbf U_N\in\mathbb R^{M\times d_N}$ with orthonormal columns
\Ensure Cleaned representations $\mathbf R\in\mathbb R^{B\times M}$

\Statex \textbf{Step 1: RFF lift and projection}
\State $\mathbf X \gets \Phi_\mathrm{rff}(\mathbf Z)$ \Comment{RFF-lifted representations from Eq.~\eqref{eq:rff}}
\State $P_{\perp} \gets \mathbf I_M - \mathbf U_N \mathbf U_N^\top$ \Comment{Projector from Eq.~\eqref{eq:proj_modality}}
\State $\mathbf R \gets \mathbf X\,P_{\perp}$ \Comment{Cleaned representations from Eq.~\eqref{eq:clean_repr}}

\Statex \textbf{Step 2: Mean-prototype update}
\For{each protected group $s = 1,\ldots,G$}
    \State $\mathcal B_s \gets \{\,i : s^{(i)} = s\,\}$
    \If{$|\mathcal B_s| > 0$}
        \State $\bar{\mathbf x}_s \gets \frac{1}{|\mathcal B_s|}\sum_{i\in\mathcal B_s}\mathbf X_{i,:}$ \Comment{Batch group mean used in Eq.~\eqref{eq:general_proto}}
        \State $\mu_s \gets (1-\rho)\,\mu_s + \rho\,\bar{\mathbf x}_s$ \Comment{EMA update from Eq.~\eqref{eq:general_proto}}
    \EndIf
\EndFor

\Statex \textbf{Step 3: SVD refresh of the prototype subspace}
\If{$t_{\mathrm{step}} \bmod T_{\mathrm{refresh}} = 0$}
    \State $\bar\mu \gets \sum_{s=1}^{G}\hat\pi_s\,\mu_s$ \Comment{Global mean for Eq.~\eqref{eq:general_nuisance_subspace}}
    \State Form $\widetilde{\mathbf P}\in\mathbb R^{G\times M}$ with rows $\widetilde{\mathbf P}_{s,:}\gets(\mu_s-\bar\mu)^\top$ \Comment{Top right singular vectors from Eq.~\eqref{eq:proto_svd}}
    \State $\mathbf U_N \gets \textsc{TopRightSingVecs}(\widetilde{\mathbf P}, d_N)$ \Comment{Centered prototype matrix for Eq.~\eqref{eq:proto_svd}}
\EndIf

\State \Return $\mathbf R$

\Statex
\Statex \textbf{Optimization step:} compute the chosen objective $\mathcal L(\mathbf R)$ and backpropagate through the RFF map $\Phi$ and the linear cleaning map $\mathbf X\mapsto \mathbf X P_{\perp}$.
\end{algorithmic}
\end{algorithm}

%%%%%%%%%%%%%%%%%%%%%%%%%%%%%%%%%%%%%%%%%%%%%%%%%%

\subsection{Supervised vs. Semi-supervised Implementation}
\label{sect:appendix_specific_implementation}

In supervised settings, the same cleaned representations \(R_m\) can be fed to %a task head 
\(h\) and trained with a standard loss (e.g., cross-entropy %for classification 
or regression loss) without modifying the \methodname block. 

%We also experiment on semi-supervised settings, where we use a two-stage protocol.
We use a two-stage protocol on semi-supervised settings.
In Stage~1 (self-supervised pretraining), we train on all available samples using either the unimodal objective \(\mathcal L_{\methodname}^{\mathrm{uni\text{-}ssl}}\) or the multimodal objective \(\mathcal L_{\methodname}^{\mathrm{mm\text{-}ssl}}\), while maintaining prototypes indexed only by \(s\) in Eq.~\eqref{eq:general_proto}.
In Stage~2 (supervised fine-tuning), we train a task head on the labeled subset using the supervised objective \(\mathcal L_{\methodname}^{\mathrm{sup}}\).
When discrete task labels are available, Stage~2 also admits a label-aware refinement in which prototypes are indexed by \((s,y)\), i.e., by replacing \(s^{(i)}\) with \((s^{(i)},y^{(i)})\) in Eq.~\eqref{eq:general_proto}. This changes only the prototypes used to form \(P_{\perp}\) and leaves the training objective unchanged. The resulting label-aware version conditions the removable nuisance structure on the task labels, reducing the risk of removing task-relevant variation in supervised classification settings.

%%%%%%%%%%%%%%%%%%%%%%%%%%%%%%%%%%%%%%%%%%%%%%%%%%

\subsection{Specific Block Formulation}
\label{sect:appendix_block_formulation}

\paragraph{Unimodal version.} 
For a mini-batch of encoder outputs
$\mathbf Z = [z^{(1)},\ldots, z^{(B)}]^\top$ with labels
$\mathbf s = (s^{(1)},\ldots,s^{(B)})$, 
we define the first-order \methodname block operator:
\begin{equation}
\begin{aligned}
\operatorname{\methodname}(\mathbf Z,\mathbf s)
&:= \Phi_\mathrm{rff}(\mathbf Z)\,P_{\perp}\\
&= \Phi_\mathrm{rff}(\mathbf Z)\bigl(I_M - U^{(N)}U^{(N)\top}\bigr).
\end{aligned}
\label{eq:fairclean-operator}
\end{equation}
where $U^{(N)}
   = \mathcal U\bigl(\{\mu_{s}\}_{s=1}^G\bigr)$ denotes the top $d_N$ right singular vectors obtained from the SVD in Eq.~\eqref{eq:proto_svd}. The map $\mathcal U$ is defined as
\begin{equation}
\begin{aligned}
\mathcal U : (\mathbb R^{M})^{G} &\to \mathbb R^{M\times d_N}\\
\{\mu_{s}\}_{s=1}^G &\longmapsto
Q_{\le1}.
\end{aligned}
\end{equation}

\paragraph{Multimodal version.}
The \methodname block could also be defined for multimodal methods that benefit from a pair of views (e.g., contrastive learning methods). For the two-view case, we define each view as $m\in\{m_1,m_2\}$. Now, we can define the multi-view version of \methodname block as
\begin{equation}
\begin{aligned}
\operatorname{\methodname}(\mathbf Z_{m_1},\mathbf Z_{m_2},\mathbf s)
&:= \bigl(\operatorname{\methodname}_{m_1}(\mathbf Z_{m_1},\mathbf s), \\
&\qquad\ \operatorname{\methodname}_{m_2}(\mathbf Z_{m_2},\mathbf s)\bigr).
\end{aligned}
\label{def:methodname}
\end{equation}

%%%%%%%%%%%%%%%%%%%%%%%%%%%%%%%%%%%%%%%%%%%%%%%%%%%%%%%%%%%%%%%%%%
%% Additional methods section details
%%%%%%%%%%%%%%%%%%%%%%%%%%%%%%%%%%%%%%%%%%%%%%%%%%%%%%%%%%%%%%%%%%

\subsection{Higher-order prototype instantiations}
\label{app:higher_order_prototypes}

Section~\ref{sect:method_prototype} defines \methodname for a general order-$K$ prototype family and introduces the first-order instantiation. Here we detail the exact second-order construction, which extends the nuisance subspace to group-conditional covariance differences in the lifted feature space.

A stronger exact second-order version additionally maintains covariance prototypes
\[
    \Sigma_s
    =
    \mathbb{E}\left[
    (X-\mu_s)(X-\mu_s)^\top \mid S=s
    \right],
    \qquad
    \bar{\Sigma}
    =
    \sum_{s=1}^{G}\pi_s\Sigma_s .
\]
The corresponding nuisance subspace is
\[
    \mathcal{N}^{\mathrm{exact}}_2
    =
    \mathrm{span}\{\mu_s-\bar{\mu}:s=1,\ldots,G\}
    +
    \sum_{s=1}^{G}\mathrm{Range}(\Sigma_s-\bar{\Sigma}).
\]

Let \(P_\perp^{(2)}\) be the orthogonal projector onto the complement of \(\mathcal N_2^{\mathrm{exact}}\). By construction, for every pair of groups \(s,t\),
\[
P_\perp^{(2)}(\mu_s-\mu_t)=0,
\qquad
P_\perp^{(2)}(\Sigma_s-\Sigma_t)P_\perp^{(2)}=0.
\]
Consequently, the cleaned representations \(R^{(2)}=P_\perp^{(2)}X\) have identical group-conditional means and covariance matrices. This extends the first-order construction from mean alignment to exact first- and second-moment alignment in the cleaned RFF space.

For orders \(q\ge 3\), forming full central moment tensors is generally impractical. We therefore use sketched higher-order prototypes based on random unit directions \(u_{q,a}\in\mathbb R^M\):
\[
\psi^{(q)}_{a}(X;\mu_s)
=
(X-\mu_s)\,
\big(u_{q,a}^\top(X-\mu_s)\big)^{q-1},
\qquad q=3,\ldots,K.
\]
These functionals provide tractable directional contractions of the group central moment tensors while keeping the prototype vectors in \(\mathbb R^M\). In implementation, the current group mini-batch mean is used for centering, and the resulting prototypes are maintained by EMA and processed through the same sequential projection framework. In the reported results, prototype order is selected by validation over orders \(1,\ldots,5\), using the task-specific validation metric described in Appendix~\ref{sect:appendix_implementation_details}.

\subsection{Two-view self-supervised loss formulations}
\label{app:loss_formulations}

Let $x_a$ and $x_b$ denote two views of the same instance, corresponding either to two augmentations of one modality or to two paired modalities. Let $f_a$ and $f_b$ be the encoders, producing $z_a=f_a(x_a)$ and $z_b=f_b(x_b)$. For group labels $s=(s^{(1)},\ldots,s^{(B)})$, the \methodname block maps each encoder output to cleaned lifted representations,
\[
    (\mathbf R_a,\mathbf R_b)
    =
    \methodname(\mathbf Z_a,\mathbf Z_b,s),
    \qquad
\mathbf R_a,\mathbf R_b\in\mathbb R^{B\times M}.
\]
Any two-view contrastive objective can then be applied without modification:
\[
    \mathcal L_{\mathrm{uni\text{-}ssl}}^{\methodname}
    =
    \mathcal L_{\mathrm{ctr}}(\mathbf R_a,\mathbf R_b),
    \qquad
    \mathcal L_{\mathrm{mm\text{-}ssl}}^{\methodname}
    =
    \mathcal L_{\mathrm{ctr}}(\mathbf R_{m_1},\mathbf R_{m_2}).
\]

%%%%%%%%%%%%%%%%%%%%%%%%%%%%%%%%%%%%%%%%%%%%%%%%%%%%%%%%%%%%%%%%%%%%%%%%%%%

\section{Detailed Theoretical Results}
\label{sect:appendix_theory}

Let \(\kappa\) be a bounded measurable kernel on the support of a cleaned
representation \(R\). For probability measures \(P,Q\) on that space, let
\(\mu_P^\kappa\) denote the corresponding kernel mean embedding, and write
\[
\mathrm{MMD}_\kappa(P,Q)
:=
\|\mu_P^\kappa-\mu_Q^\kappa\|_{\mathcal H_\kappa}.
\]
When \(\kappa\) is characteristic, \(\mathrm{MMD}_\kappa(P,Q)=0\) is equivalent
to \(P=Q\) \cite{gretton2012kernel,muandet2017kernel,sriperumbudur2011universality}.

\begin{definition}[Shared-base affine structural model]
\label{def:affine_world}
Take a generic encoder representation \(Z\in\mathbb R^d\), a discrete protected
attribute \(S\in\{1,\dots,G\}\) with \(\pi_g:=\Pr(S=g)>0\) for every \(g\), and
a latent random vector \(W\in\mathbb R^d\) satisfying
\[
W\perp S.
\]
For each group \(g\in\{1,\dots,G\}\), let
\(A_g\in\mathbb R^{d\times d}\) and \(b_g\in\mathbb R^d\), and write
\[
Z=A_S W+b_S
\qquad \text{almost surely.}
\]
Let
\[
\mathcal H:=\mathrm{span}\bigl(\mathrm{supp}(W)\bigr)\subseteq\mathbb R^d
\]
denote the linear span of the support of \(W\).
\end{definition}

\begin{definition}[Canonical affine discrepancy subspace]
\label{def:canonical_nuisance}
In the setting of Definition~\ref{def:affine_world}, define
\[
\mathcal N_{\mathrm{aff}}
:=
\mathrm{span}\!\Big(
\{\,b_g-b_h : g,h\in\{1,\dots,G\}\,\}
\;\cup\;
\bigcup_{g,h\in\{1,\dots,G\}} (A_g-A_h)\mathcal H
\Big).
\]
Equivalently,
\[
\mathcal N_{\mathrm{aff}}
=
\mathrm{span}\!\left\{
(A_g-A_h)w+(b_g-b_h)
:
g,h\in\{1,\dots,G\},\; w\in\mathcal H
\right\}.
\]
Let \(P_{\mathrm{aff},\perp}:\mathbb R^d\to\mathcal N_{\mathrm{aff}}^\perp\)
denote the orthogonal projector onto \(\mathcal N_{\mathrm{aff}}^\perp\), and define
\[
R^\star:=P_{\mathrm{aff},\perp}Z.
\]
\end{definition}

Definition~\ref{def:canonical_nuisance} collects exactly the linear directions
along which two group-specific affine images of the same latent point can differ.

\begin{theorem}[Canonical affine projection theorem]
\label{thm:canonical_affine_projection}
In the setting of Definition~\ref{def:affine_world}, there exists a measurable map
\[
T:\mathcal H\to\mathcal N_{\mathrm{aff}}^\perp
\]
such that
\[
R^\star=T(W)
\qquad \text{almost surely.}
\]
Consequently,
\[
R^\star\perp S.
\]

Moreover, if \(\mathcal M\subseteq\mathbb R^d\) is a linear subspace satisfying
\[
P_{\mathcal M^\perp}(A_gw+b_g)
=
P_{\mathcal M^\perp}(A_hw+b_h)
\qquad
\text{for every } g,h\in\{1,\dots,G\},\; w\in\mathcal H,
\]
then
\[
\mathcal N_{\mathrm{aff}}\subseteq \mathcal M.
\]
Hence \(\mathcal N_{\mathrm{aff}}\) is the smallest linear subspace whose
removal equalizes all groups under the affine structural model.
\end{theorem}

\begin{proof}
For every \(g,h\in\{1,\dots,G\}\) and \(w\in\mathcal H\),
\[
(A_gw+b_g)-(A_hw+b_h)
=
(A_g-A_h)w+(b_g-b_h)\in\mathcal N_{\mathrm{aff}}.
\]
Therefore
\[
P_{\mathrm{aff},\perp}(A_gw+b_g)
=
P_{\mathrm{aff},\perp}(A_hw+b_h).
\]
Fix a reference group \(g_0\in\{1,\dots,G\}\), and define
\[
T(w):=P_{\mathrm{aff},\perp}(A_{g_0}w+b_{g_0}),
\qquad w\in\mathcal H.
\]
The preceding identity shows that this quantity does not depend on the chosen
group after projection, so
\[
P_{\mathrm{aff},\perp}(A_gw+b_g)=T(w)
\qquad
\text{for every } g\in\{1,\dots,G\},\; w\in\mathcal H.
\]
Since \(W\in\mathcal H\) almost surely and \(Z=A_SW+b_S\) almost surely,
\[
R^\star
=
P_{\mathrm{aff},\perp}Z
=
P_{\mathrm{aff},\perp}(A_SW+b_S)
=
T(W)
\qquad \text{almost surely.}
\]
Because \(W\perp S\), it follows that
\[
R^\star\perp S.
\]

For minimality, let \(\mathcal M\subseteq\mathbb R^d\) satisfy the stated equalization
property. Then
\[
P_{\mathcal M^\perp}\bigl((A_g-A_h)w+(b_g-b_h)\bigr)=0
\qquad
\text{for every } g,h\in\{1,\dots,G\},\; w\in\mathcal H,
\]
hence
\[
(A_g-A_h)w+(b_g-b_h)\in\mathcal M
\qquad
\text{for every } g,h,w.
\]
Taking \(w=0\) gives \(b_g-b_h\in\mathcal M\). Subtracting those vectors then gives
\[
(A_g-A_h)w\in\mathcal M
\qquad
\text{for every } g,h\in\{1,\dots,G\},\; w\in\mathcal H.
\]
Thus \(\mathcal M\) contains all generators of \(\mathcal N_{\mathrm{aff}}\), so
\[
\mathcal N_{\mathrm{aff}}\subseteq \mathcal M.
\]
\end{proof}

Theorem~\ref{thm:canonical_affine_projection} establishes the geometric basis for removing group-dependent directions under the shared-base affine model. We next analyze the prototype-based projection used by \methodname, directly through the group-conditioned statistics defined in Section~\ref{sect:method_prototype}.

\begin{definition}[KINDER prototype-space invariance]
\label{def:prototype_family}
Fix the order-$K$ prototype family $\mathcal F_K$ from Section~\ref{sect:method_prototype}, and let $X=\Phi_{\mathrm{rff}}(Z)\in\mathbb R^M$. For the population analysis, form $\mathcal N_K$ as in Eq.~\eqref{eq:general_nuisance_subspace}, using population conditional expectations in place of EMA prototypes and population group frequencies in place of empirical frequencies. Let $P_\perp^{(K)}$ project onto $\mathcal N_K^\perp$, with all nonzero nuisance directions retained, and set
\[
R=P_\perp^{(K)}X.
\]
For each selected order $q$, index $a\in\mathcal A_q$, and protected group $s$, define the population prototype
\[
m_{s,a}^{(q)}:=\mathbb E[\psi_a^{(q)}(X;\mu_s)\mid S=s].
\]
We write $R\perp_{\mathcal F_K}S$ when the selected projected prototypes agree across groups:
\[
P_\perp^{(K)}m_{g,a}^{(q)}=P_\perp^{(K)}m_{h,a}^{(q)}
\quad\text{for every }g,h,\ q=1,\ldots,K,\ a\in\mathcal A_q.
\]
Equivalently, each selected pair of projected prototype distributions has zero MMD under the linear kernel $\ell(u,v)=u^\top v$ on $\mathbb R^M$. For first-order prototypes, this is zero pairwise linear-kernel MMD between the group-conditional laws of $R$; higher orders control their corresponding selected projected prototype statistics.
\end{definition}

\begin{theorem}[\methodname\ prototype-space independence theorem]
\label{thm:kinder_f_independence}
In the setting of Definition~\ref{def:prototype_family},
\[
R\perp_{\mathcal F_K}S.
\]
\end{theorem}

\begin{proof}
For every selected $q,a$, the population prototype differences $m_{s,a}^{(q)}-\sum_t\pi_t m_{t,a}^{(q)}$ belong to $\mathcal N_K$ by construction. Hence their projections vanish, and
\[
P_\perp^{(K)}m_{g,a}^{(q)}=P_\perp^{(K)}m_{h,a}^{(q)}
\qquad\text{for all }g,h,q,a.
\]
These quantities are the conditional means of the projected prototype vectors $P_\perp^{(K)}\psi_a^{(q)}(X;\mu_s)$. Their equality gives zero pairwise MMD under the linear kernel $\ell$ for every selected prototype, proving $R\perp_{\mathcal F_K}S$.
\end{proof}

Theorem~\ref{thm:kinder_f_independence} establishes the exact population prototype-space guarantee for the selected family. Its first-order case matches group-conditional means of the cleaned representation. The exact second-order covariance construction is detailed in Appendix~\ref{app:higher_order_prototypes}.

\begin{corollary}[Exact independence under law-identifying prototypes]
\label{cor:exact_independence}
In the setting of Theorem~\ref{thm:kinder_f_independence}, let $\mathcal P_{\mathcal F_K}$ be a class of laws on the lifted feature space containing each $P_{X\mid S=s}$. Suppose the selected projected prototypes identify the cleaned laws on this class. Meaning, for any $P,Q\in\mathcal P_{\mathcal F_K}$ with $\mu_P=\mathbb E_{X\sim P}[X]$ and $\mu_Q=\mathbb E_{X\sim Q}[X]$, assume that
\[
P_\perp^{(K)}\mathbb E_{X\sim P}[\psi_a^{(q)}(X;\mu_P)]
=P_\perp^{(K)}\mathbb E_{X\sim Q}[\psi_a^{(q)}(X;\mu_Q)]
\quad\text{for all }q,a
\]
implies equality of the laws of $P_\perp^{(K)}X$ under $P$ and $Q$. Then
\[
R\perp S.
\]
\end{corollary}

\begin{proof}
Theorem~\ref{thm:kinder_f_independence} equalizes all selected projected prototypes of the group-conditional laws $P_{X\mid S=g}$ and $P_{X\mid S=h}$. The law-identifying property therefore makes the corresponding cleaned laws $P_{R\mid S=g}$ and $P_{R\mid S=h}$ equal for every pair $g,h$. Since $S$ is discrete, these equal conditional laws give $R\perp S$.
\end{proof}

Theorem~\ref{thm:kinder_f_independence} describes the population projector
for a fixed RFF map. We now quantify the effect of estimating its nuisance
subspace and retaining a finite number of singular directions.

For each order $q$, let $\widetilde P_q$ be the population centered
prototype matrix from Section~\ref{sect:method_prototype}, formed from
the conditional means $m_{s,a}^{(q)}$ in
Definition~\ref{def:prototype_family}. Let
$\widehat{\widetilde P}_q$ be its EMA-based estimate, including the
empirical group frequencies. Write $\Pi_{\le q}$ and
$\widehat\Pi_{\le q}$ for the cumulative population and empirical
nuisance projectors constructed by the sequential SVD in
Eqs.~\eqref{eq:proto_svd}--\eqref{eq:proj_modality}, with
$\Pi_{\le0}=\widehat\Pi_{\le0}=0$.

\begin{proposition}[Finite-sample prototype-space discrepancy]
\label{prop:finite_sample_prototype}
At each order $q$, let
\[
A_q=\widetilde P_q(I_M-\Pi_{<q}),\qquad
\widehat A_q=\widehat{\widetilde P}_q(I_M-\widehat\Pi_{<q}),
\]
where $\Pi_{<q}=\Pi_{\le q-1}$ and similarly for its estimate.
Suppose $r_q$ nonzero right-singular directions are selected at
order $q$. Define
\[
\epsilon_q=\|\widehat{\widetilde P}_q-\widetilde P_q\|_{\mathrm{op}},
\quad
\Delta_q=\sigma_{r_q}(A_q)-\sigma_{r_q+1}(A_q).
\]
Set $b_0=0$ and, recursively,
\[
\zeta_q=\epsilon_q+\|\widehat{\widetilde P}_q\|_{\mathrm{op}}b_{q-1},
\qquad
b_q=b_{q-1}+\frac{4\zeta_q}{\Delta_q-\zeta_q}.
\]
If $\Delta_q>2\zeta_q$ for every selected order, then
\[
\|\widehat\Pi_{\le K}-\Pi_{\le K}\|_{\mathrm{op}}\le b_K.
\]
Consequently, for every group pair $g,h$ and selected prototype $q,a$,
writing $d_{g,h}^{(q,a)}=m_{g,a}^{(q)}-m_{h,a}^{(q)}$,
\begin{equation}
\begin{aligned}
\bigl\|(I_M-\widehat\Pi_{\le K})d_{g,h}^{(q,a)}\bigr\|_2
&\le\underbrace{\bigl\|(I_M-\Pi_{\le K})d_{g,h}^{(q,a)}\bigr\|_2}_{\text{rank-truncation residual}}\\
&\quad+b_K\bigl\|d_{g,h}^{(q,a)}\bigr\|_2.
\end{aligned}
\label{eq:finite_sample_mmd}
\end{equation}
The left-hand side is the pairwise linear-kernel MMD between the
corresponding projected group-conditional prototype distributions.
When all nonzero nuisance directions are retained, the first term
vanishes.
\end{proposition}

\begin{proof}
Let $\Pi_q=\Pi_{\le q}-\Pi_{<q}$ be the projector onto the
new directions selected at order $q$, and define $\widehat\Pi_q$
analogously. The residualized matrices satisfy
\[
\|\widehat A_q-A_q\|_{\mathrm{op}}
\le\epsilon_q+\|\widehat{\widetilde P}_q\|_{\mathrm{op}}
\|\widehat\Pi_{<q}-\Pi_{<q}\|_{\mathrm{op}}.
\]
By induction, this difference is at most $\zeta_q$. Wedin's
singular-subspace perturbation inequality and
$\Delta_q>2\zeta_q$ give the conservative estimate
\[
\|\widehat\Pi_q-\Pi_q\|_{\mathrm{op}}
\le\frac{4\zeta_q}{\Delta_q-\zeta_q}.
\]
Because the selected directions are orthogonal across orders,
$\Pi_{\le q}=\sum_{j=1}^q\Pi_j$ and
$\widehat\Pi_{\le q}=\sum_{j=1}^q\widehat\Pi_j$.
The triangle inequality therefore yields
$\|\widehat\Pi_{\le q}-\Pi_{\le q}\|_{\mathrm{op}}\le b_q$.
Applying it again to
$(I_M-\widehat\Pi_{\le K})d_{g,h}^{(q,a)}$ proves
Eq.~\eqref{eq:finite_sample_mmd}. For the linear kernel, MMD is
the norm of the conditional mean difference. If all nonzero
singular directions are selected at every order, the cumulative
population subspace is $\mathcal N_K$, which contains each
$d_{g,h}^{(q,a)}$.
\end{proof}

The perturbations $\epsilon_q$ can be related to group sample sizes.
Suppose all selected prototype vectors have norm at most $B$, and put
\[
e=\max_{s,q,a}\|p_{s,a}^{(q)}-m_{s,a}^{(q)}\|_2,
\qquad
\nu=\sum_{s=1}^G|\widehat\pi_s-\pi_s|.
\]
The weighted centering step gives, for each order,
\begin{equation}
\epsilon_q\le\sqrt{G|\mathcal A_q|}\,(2e+B\nu).
\label{eq:prototype_estimation_error}
\end{equation}
For a fixed encoder and RFF map, suppose each group provides at least
$n_{\min}$ independent observations. If the EMA and evolving-feature
tracking error relative to the corresponding empirical prototype means
is at most $\beta$, bounded-vector concentration gives, with
probability at least $1-\alpha$,
\begin{equation}
e\le\beta+4B\sqrt{\frac{\log(2J/\alpha)}{n_{\min}}},
\qquad J=G\sum_{q=1}^K|\mathcal A_q|.
\label{eq:prototype_concentration}
\end{equation}
Thus, subgroup size controls the sampling term, while $\beta$ and
$\nu$ account for EMA tracking and empirical group-frequency error.

\subsection{Random Fourier feature approximation}
\label{app:rff_approximation}

The preceding result is conditional on the realized finite RFF map.
The following bound quantifies its approximation of the Gaussian
kernel used in Section~\ref{sect:method_lifting}.

\begin{proposition}[Gaussian-to-RFF kernel approximation]
\label{prop:rff_approximation}
Let $k_Z$ be the Gaussian kernel defined in Section~\ref{sect:method_lifting},
and define
\[
k_M(z,z')=\langle\Phi_{\mathrm{rff}}(z),
                 \Phi_{\mathrm{rff}}(z')\rangle.
\]
On a compact domain $\mathcal Z\subset\mathbb R^d$, put
\[
\varepsilon_M=
\sup_{z,z'\in\mathcal Z}|k_M(z,z')-k_Z(z,z')|.
\]
For any probability distributions $P,Q$ supported on $\mathcal Z$,
\begin{equation}
\left|\mathrm{MMD}_{k_M}^2(P,Q)
-\mathrm{MMD}_{k_Z}^2(P,Q)\right|\leq4\varepsilon_M,
\label{eq:rff_mmd_approximation}
\end{equation}
and therefore
\[
\left|\mathrm{MMD}_{k_M}(P,Q)
-\mathrm{MMD}_{k_Z}(P,Q)\right|\leq2\sqrt{\varepsilon_M}.
\]
For any fixed $z,z'$, the RFF construction in Eq.~\eqref{eq:rff}
also satisfies
\[
\Pr\!\left( |k_M(z,z')-k_Z(z,z')|>\varepsilon\right)
\leq2\exp\!\left(-\frac{M\varepsilon^2}{8}\right).
\]
\end{proposition}

\begin{proof}
The squared MMD is the sum of within-distribution kernel expectations
minus twice the cross-distribution kernel expectation. Each changes by
at most $\varepsilon_M$ when $k_Z$ is replaced by $k_M$, yielding
Eq.~\eqref{eq:rff_mmd_approximation}. The unsquared bound follows from
$|\sqrt a-\sqrt b|\leq\sqrt{|a-b|}$ for $a,b\geq0$.
For fixed $z,z'$, the RFF kernel is an average of $M$ independent,
unbiased terms of the form
$2\cos(\omega_j^\top z+b_j)\cos(\omega_j^\top z'+b_j)$,
each in $[-2,2]$. Hoeffding's inequality gives the stated tail bound.
\end{proof}

Proposition~\ref{prop:rff_approximation} bounds approximation of the
base Gaussian kernel by the finite RFF map. For that realized map,
Theorem~\ref{thm:kinder_f_independence} gives exact population
prototype-space invariance when the full nuisance subspace is removed;
Proposition~\ref{prop:finite_sample_prototype} quantifies the additional
residual from rank truncation and empirical prototype estimation.

\section{More on Experimental Settings}
\label{sect:appendix_experimental_setting}

\subsection{Datasets}
\label{sect:appendix_datasets}

\paragraph{Bias for Bios.}
We use BIOS for occupation classification from free-text biographies. BIOS was introduced to study bias in occupation prediction from semantic representations, and is particularly well suited for concept-erasure evaluation because the target task is strongly correlated with demographic cues such as gender.  \cite{de2019bias}

\paragraph{Moji.}
We use Moji for binary sentiment classification from tweets. The dataset is designed to study social bias in language representations by comparing sentiment prediction across dialect groups. The protected attribute is dialect, distinguishing African-American English from Standard American English. This setting is useful for evaluating whether a model’s sentiment predictions rely on dialectal cues rather than task-relevant semantic content. 
\cite{blodgett2016demographic}

\paragraph{Harvard-FairVLMed.}
We use Harvard-FairVLMed for multimodal vision-language learning. The task is binary glaucoma classification from paired Scanning Laser Ophthalmoscopy fundus images and de-identified ophthalmology clinical notes. The dataset contains 10{,}000 image--text pairs with official training, validation, and test splits. For fairness evaluation, we use race, gender, age group, and ethnicity as protected attributes
\cite{luo2024fairclip}.

\paragraph{MIMIC-IV.}
We use MIMIC-IV for both classification and regression experiments. MIMIC-IV is a large deidentified electronic health record database collected from Beth Israel Deaconess Medical Center and contains rich clinical information including demographics, diagnoses, procedures, laboratory measurements, treatments, and notes. Its breadth makes it a useful benchmark for studying fairness in realistic clinical prediction settings.
\cite{johnson2020mimic}

\paragraph{FairSeg.}
We evaluate segmentation on FairSeg, a medical-image fairness benchmark built from 10{,}000 fundus images with pixel-level disc and cup masks and associated demographic attributes. FairSeg was introduced to support fairness research in medical segmentation, where both predictive quality and group-level equity are important.
\cite{tianfairseg}

\paragraph{CelebA.}
We use CelebA for unimodal face-attribute prediction. The task is binary smile classification from face images. Images and attribute labels are taken from the original CelebA dataset, using its standard train/validation/test partition. For fairness evaluation, we use age group and gender as protected attributes~\cite{liu2015deep}.

% \paragraph{ACSIncome.}
% We evaluate tabular classification on ACSIncome, a Folktables task derived from the American Community Survey and designed as a larger and more recent replacement for the widely used UCI Adult benchmark. The task is binary income prediction, and the dataset is commonly used to study fairness under demographic attributes such as gender and race.
% \cite{ding2021retiring}

\paragraph{Communities and Crime.}
We use Communities and Crime for tabular regression. This dataset combines community-level socioeconomic variables, law-enforcement statistics, and crime outcomes across U.S.\ communities, making it a standard benchmark for studying regression under sensitive demographic structure.
\cite{mcclendon2015using}

%%%%%%%%%%%%%%%%%%%%%%%
%%%%%%%%%%%%%%%%%%%%%%%

\paragraph{Dataset split imbalance.}
Table~\ref{tab:dataset_split_imbalance} summarizes the imbalance structure of the evaluation splits under the same protected attributes used in the reported fairness metrics. 
The table separates marginal protected-group imbalance from the more difficult label-conditioned or target-conditioned subgroup imbalance. 
This distinction is important because several datasets that appear moderate under a single protected attribute become substantially more skewed after conditioning on the task label or on protected-attribute intersections. 
The table should be read as a characterization of the statistical stress imposed by the benchmark: classification settings test whether errors and true-positive behavior remain stable across label-conditioned protected groups; segmentation settings test whether spatial prediction quality remains stable across protected groups and their intersections; and regression settings test whether prediction error remains stable across groups with different target distributions.

These split properties are closely related to the design of \methodname. 
Because \methodname estimates protected-group structure through group prototypes in the representation space, it is intended to remove demographic variation without tying the erasure operation to a particular prediction head or task loss. 
The imbalance summary therefore helps explain why the experiments emphasize not only marginal protected attributes, but also multi-attribute and intersectional settings. In these regimes, fairness failures can remain hidden when only aggregate performance or single-attribute summaries are considered. 
The empirical results are broadly consistent with this motivation, especially in the text, MIMIC-IV classification, FairSeg, and FairVLMed settings, where improvements are observed on fairness metrics that directly depend on class-conditioned, multi-attribute, or intersectional subgroup behavior. 
At the same time, the regression results indicate that severe intersectional sparsity remains a difficult case.

\paragraph{Dataset and asset licenses.}
All datasets and baseline assets are used according to their original research-use terms. MIMIC-IV is used under the PhysioNet credentialed-access data use agreement. CelebA is used under its non-commercial research-use terms. BIOS, Moji, FairSeg, Harvard-FairVLMed, and Communities and Crime are public research datasets used according to their original release terms. We do not redistribute any dataset.

\begin{table}[t]
\centering
\caption{Dataset split imbalance summary. 
$R$ denotes the max/min ratio over nonzero groups or cells. 
For classification datasets, the final column reports the smallest nonzero label-by-intersectional protected cell. 
For segmentation, it reports the smallest intersectional protected group. 
For regression, it reports the largest target-mean shift across protected groups or intersections. 
Non-specific protected labels such as unknown/other are excluded.}
\label{tab:dataset_split_imbalance}

\scriptsize
\setlength{\tabcolsep}{1pt}
\renewcommand{\arraystretch}{1.24}
\newcommand{\tcell}[1]{\begin{tabular}[t]{@{}l@{}}#1\end{tabular}}

\begin{tabular}{@{}
p{0.095\textwidth}
p{0.135\textwidth}
l
p{0.105\textwidth}
p{0.360\textwidth}
p{0.245\textwidth}
@{}}
\toprule
\textbf{Dataset} &
\textbf{Task / target} &
\textbf{$N$} &
\textbf{Outcome} &
\textbf{Smallest group per attribute} &
\textbf{Worst split imbalance} \\
\midrule

% CelebA &
% \tcell{Binary\\classification\\Smiling} &
% 202{,}599 &
% positive 48.2\% &
% \tcell{
% gender: male, 41.7\%, $R{=}1.4{\times}$\\
% age: not-young, 22.6\%, $R{=}3.4{\times}$
% } &
% \tcell{negative $\times$ not-male\\$\times$ not-young\\2.3\%, $R{=}11.5{\times}$} \\

FairSeg &
\tcell{Segmentation\\disc/cup mask} &
9{,}432 &
mask target &
\tcell{
race: Asian, 9.2\%, 
$R{=}8.3{\times}$\\
gender: male, 41.5\%, $R{=}1.4{\times}$\\
ethnicity: Hispanic, 3.9\%, $R{=}24.4{\times}$
} &
\tcell{Asian $\times$ male\\$\times$ Hispanic\\$<0.1$\%, $R{=}4051.0{\times}$} \\

FairVLMed &
\tcell{Binary\\classification\\glaucoma} &
9{,}460 &
positive 50.6\% &
\tcell{
race: Asian, 8.3\%, $R{=}9.3{\times}$\\
gender: male, 43.6\%, $R{=}1.3{\times}$\\
ethnicity: Hispanic, 4.2\%, $R{=}22.8{\times}$
} &
\tcell{positive $\times$ Asian\\$\times$ male $\times$ Hispanic\\$<0.1$\%, $R{=}2089.0{\times}$} \\

MIMIC-IV ICU mortality &
\tcell{Binary\\classification\\hospital mortality} &
72{,}114 &
positive 10.1\% &
\tcell{
gender: female, 45.0\%, $R{=}1.2{\times}$\\
race: NHPI, 0.2\%, $R{=}483.3{\times}$\\
age: $\leq$50, 16.4\%, $R{=}5.1{\times}$
} &
\tcell{positive $\times$ female\\$\times$ AIAN $\times$ $\leq$50\\$<0.1$\%, $R{=}24464.0{\times}$} \\

MIMIC-IV ICU LOS &
\tcell{Regression\\ICU length\\of stay} &
56{,}527 &
\tcell{median 2.36\\IQR 1.55--4.29} &
\tcell{
gender: female, 44.2\%, $R{=}1.3{\times}$\\
race: NHPI, 0.2\%, $R{=}524.8{\times}$\\
age: 18--39, 7.9\%, $R{=}4.5{\times}$
} &
\tcell{intersectional\\target-mean shift\\5.8} \\

Bias for Bios &
\tcell{Multiclass\\classification\\profession} &
396{,}189 &
\tcell{28 classes\\class $R{=}84.2{\times}$} &
\tcell{gender: group 1, 46.1\%, $R{=}1.2{\times}$} &
\tcell{profession 23\\$\times$ group 1\\$<0.1$\%, $R{=}476.6{\times}$} \\

Moji 90\% &
\tcell{Binary\\classification\\sentiment} &
100{,}000 &
positive 50.0\% &
\tcell{
dialect: protected=0, 50.0\%, $R{=}1.0{\times}$
} &
\tcell{negative $\times$ protected=1\\5.0\%, $R{=}9.0{\times}$} \\

Communities and Crime &
\tcell{Regression\\violent crime\\rate} &
1{,}994 &
\tcell{median 0.15\\IQR 0.07--0.33} &
\tcell{
race: Asian, 4.4\%, $R{=}17.9{\times}$
} &
\tcell{race target-mean shift\\0.392} \\

\bottomrule
\end{tabular}
\end{table}

%%%%%%%%%%%%%%%%%%%%%%%
%%%%%%%%%%%%%%%%%%%%%%%

\subsection{Implementation Details}
\label{sect:appendix_implementation_details}
\noindent\textbf{Software and optimization.}
Unless specified otherwise by a baseline paper, all models were implemented in PyTorch. Models were optimized with Adam~\cite{kingma2014adam} or AdamW~\cite{loshchilov2017adamw}, matching the optimizer used by the corresponding baseline trainer when applicable. Hyperparameters were selected according to validation task utility, using AUC for classification, MAE for regression, and Dice score for segmentation.

For all experiments, the main hyperparameter tuned for \methodname{} was the Gamma parameter, searched over $\Gamma \in \{0.1, 0.5, 1.0, 1.5, 2.0, 10, 15, 50\}$. We also performed a smaller grid search over the random Fourier feature dimension, $m_{\mathrm{RFF}} \in \{128, 256, 512, 1024\}$. All remaining hyperparameters were kept identical to those of the corresponding baseline model.

\paragraph{Compute resources.}
All experiments were run on a cluster using a single NVIDIA A100 80GB GPU per run. Individual training runs usually took up to 3 GPU-hours depending on the task, dataset, and backbone. 
Hyperparameter grid searches required approximately 1--48 GPU-hours depending on the size of the search space and the task.

\begin{table}[t]
\centering
\caption{\textbf{MIMIC-IV classification} computational-cost comparison. Method fitting time measures the erasure stage. Total pipeline time includes backbone training, representation extraction for post-hoc methods, method fitting, validation search, and downstream-head fitting. Peak GPU memory and RAM are measured over the complete pipeline.}
\label{tab:rebuttal_compute_cost}
\small
\resizebox{\columnwidth}{!}{%
\begin{tabular}{lcccc}
\toprule
\textbf{Method} & \textbf{Method fitting (s) $\downarrow$} & \textbf{Total pipeline (s) $\downarrow$} & \textbf{Peak GPU (GB) $\downarrow$} & \textbf{Peak RAM (GB) $\downarrow$} \\
\midrule
TabM & -- & 21.6 & 0.147 & 1.832 \\
INLP & 171.7 & 195.1 & 0.147 & 3.654 \\
LEACE & 1.4 & 24.8 & 0.147 & 3.129 \\
SAL & 7.2 & 30.2 & 0.147 & 3.943 \\
KCE & 150.1 & 173.2 & 0.682 & 6.721 \\
RLACE & 217.4 & 240.8 & 0.147 & 5.677 \\
\cmidrule(lr){1-5}
\methodname-order 1 & 2.9 & 25.6 & 0.147 & 2.876 \\
\methodname-order 2 & 7.9 & 30.2 & 0.147 & 2.885 \\
\methodname-order 3 & 9.0 & 31.3 & 0.147 & 2.902 \\
\bottomrule
\end{tabular}
}
\end{table}

\noindent\textbf{Self-supervised learning experiments.}
For SSL methods, we swept a $3 \times 3$ grid over learning rates $\{10^{-5}, 10^{-4}, 10^{-3}\}$ and weight-decay coefficients $\{10^{-6.5}, 10^{-5}, 10^{-3.5}\}$. These values correspond to three log-spaced points spanning the same decade ranges as in prior work~\cite{chen2020simple}. Objective-specific architectural hyperparameters were fixed to their default values for each loss and architecture. Within this grid, SSL models were trained for $10$ epochs. We used a batch size of $512$ for multimodal runs and $1024$ for unimodal runs.

\noindent\textbf{Text classification experiments.}
For text classification, we used a batch size of $512$, a learning rate of $2 \times 10^{-5}$, a weight decay of $10^{-2}$, and early stopping patience of $3$. These hyperparameters were kept fixed across the corresponding baseline and \methodname{} experiments.

\noindent\textbf{Regression experiments.}
For regression tasks, we used a batch size of $1024$, an evaluation batch size of $4096$, and a maximum training length of $50$ epochs. Model selection was performed using validation MAE, and the same optimization and model-selection protocol was used for both the baseline and \methodname{} experiments.

\noindent\textbf{Segmentation experiments.}
For segmentation tasks, we used a batch size of $64$, a learning rate of $10^{-3}$, a weight decay of $10^{-4}$, and early stopping patience of $3$. These hyperparameters were kept fixed across all segmentation experiments. Importantly, prior concept-erasure methods were not directly compatible with segmentation tasks, since they typically require removing or retraining the higher-level learned weights after the layer to which the erasure method is applied. To adapt these methods to segmentation, we used the following protocol: first, we trained the baseline model; second, we applied the concept-erasure method to the encoder; and third, we retrained the decoder on the cleaned representations while keeping the encoder frozen.
% learning rate of ijepa and lejepa changed from $10^{-3}$ to $10^{-5}$
% comments applied right above

\noindent\textbf{Representation learning backbones.}
% (old version) Across benchmarks, we list only backbones that are reported in the paper tables. For multimodal clinical experiments on Harvard-FairVLMed and MIMIC-CXR, we use CLIP ViT-B/32 image and text encoders \cite{radford2021learning,luo2024fairclip}, with ResNet-50 \cite{he2016deep} as an alternative image backbone in the reported ablations. For CelebA-HQ and FairDomain unimodal image SSL, we report CLIP ViT-B/32 \cite{radford2021learning} and ResNet-50 \cite{he2016deep}. For MIMIC-IV tabular SSL, we report an MLP baseline together with tabular ResNet and FT-Transformer backbones \cite{gorishniy2021revisiting}. For ACSIncome and Communities \& Crime supervised tabular runs, we report periodic-embedding MLPs (MLP-PLR) \cite{gorishniy2021revisiting}. For Bias in Bios and Moji, we report Transformer text encoders from the HuggingFace ecosystem \cite{wolf2020transformers}, with tokenization and fine-tuning schedules following the model-specific defaults used in our experiments. For FairSeg, we follow the backbone and checkpoint choices reported in the FairSeg benchmark comparisons \cite{tianfairseg}. Protected attributes are used only in fairness evaluation and in the prototype-indexed \methodname variants.
For Harvard-FairVLMed multimodal experiments, we report CLIP and FactorCL backbones from the FairVLMed/FairCLIP setting \cite{radford2021learning,luo2024fairclip}. For MIMIC-IV tabular \emph{classification}, the backbone is TabM \cite{gorishniy2024tabm}. For tabular \emph{regression}, we report ResNet \cite{he2016deep} and periodic-embedding MLPs (MLP-PLR) \cite{gorishniy2021revisiting} on MIMIC-IV, and MLP-PLR on Communities and Crime. For Bias in Bios and Moji, we use the BERT Transformer encoder from HuggingFace \cite{wolf2020transformers}. For FairSeg, we use U-Net \cite{ronneberger2015u} following the reported FairSeg benchmark configuration \cite{tianfairseg}. Protected attributes are used only in fairness evaluation and in the prototype-indexed \methodname variants.

\subsection{Compared Method Details}
\label{sect:appendix_compared_method_details}
\noindent We compare \methodname-augmented training against standard post-training \emph{concept erasure} and projection baselines applied to frozen features before probing. The compared baselines are INLP \cite{ravfogel2020null}, RLACE \cite{ravfogel2022linear}, KCE \cite{ravfogel2022adversarial}, LEACE \cite{belrose2023leace}, and SUP \cite{shi2024debiasing}. For each method, we follow the implementation and optimization details reported in the corresponding paper whenever applicable, and tune only the main erasure-strength hyperparameter used by that method: the projection count for INLP, the removed dimension for RLACE, the kernel family and kernel bandwidth $\gamma$ for KCE, the sufficient-projection dimension $q$ for SUP, and the removed rank for LEACE. All baseline hyperparameters are selected by validation task utility rather than fairness or erasure scores, using the task-specific validation criteria described in Appendix~\ref{sect:appendix_implementation_details}. This matches our research question on how much protected-attribute information can be removed while preserving strong downstream performance.

\subsection{Definitions of Metrics Used}
\label{sect:appendix_fairness_measure_definiton}

Let $\mathcal{D}=\{(x_i,y_i,\hat{y}_i,s_i)\}_{i=1}^{n}$ denote the test set, where
$y_i$ is the ground-truth target, $\hat{y}_i$ is the model prediction, and
$s_i \in \mathcal{G}$ denotes the protected-group membership. The group variable
$s_i$ may denote either a single protected attribute or an intersectional group
obtained by taking the Cartesian product of multiple protected attributes. Thus,
all group-wise metrics below are computed in the same way for both marginal and
intersectional fairness evaluations.

\label{sec:metrics}

Let $Y$ denote the ground-truth target, $\hat{Y}$ the model prediction, and
$S \in \mathcal{G}$ the protected-group variable. The group variable $S$ can
represent either a single protected attribute or an intersectional attribute
formed from multiple protected attributes; all group-wise metrics are computed
with respect to the resulting group set $\mathcal{G}$.

\paragraph{Classification.}
For non-text classification tasks, we follow prior work \citep{cheong2025fairwell}
and report an aggregate fairness metric based on statistical parity, equal
opportunity, equalized odds, and equal accuracy. Let
\begin{align}
\mathrm{SR}(g) &= \mathbb{P}(\hat{Y}=1 \mid S=g), \\
\mathrm{TPR}(g) &= \mathbb{P}(\hat{Y}=1 \mid Y=1,S=g), \\
\mathrm{FPR}(g) &= \mathbb{P}(\hat{Y}=1 \mid Y=0,S=g), \\
\mathrm{Acc}(g) &= \mathbb{P}(\hat{Y}=Y \mid S=g).
\end{align}
For a group-wise quantity $m(g)$, define its group disparity as
\begin{equation}
\Delta_{\mathcal{G}}(m)
=
\max_{g\in\mathcal{G}} m(g)
-
\min_{g\in\mathcal{G}} m(g).
\end{equation}
We then compute
\begin{align}
\mathrm{SP}_{\mathrm{gap}}
&=
\Delta_{\mathcal{G}}(\mathrm{SR}), \\
\mathrm{EOpp}_{\mathrm{gap}}
&=
\Delta_{\mathcal{G}}(\mathrm{TPR}), \\
\mathrm{EOdd}_{\mathrm{gap}}
&=
\max\left\{
\Delta_{\mathcal{G}}(\mathrm{TPR}),
\Delta_{\mathcal{G}}(\mathrm{FPR})
\right\}, \\
\mathrm{EAcc}_{\mathrm{gap}}
&=
\Delta_{\mathcal{G}}(\mathrm{Acc}).
\end{align}
The aggregate fairness gap is
\begin{equation}
\mathrm{FairGap}
=
\frac{1}{4}
\left(
\mathrm{SP}_{\mathrm{gap}}
+
\mathrm{EOpp}_{\mathrm{gap}}
+
\mathrm{EOdd}_{\mathrm{gap}}
+
\mathrm{EAcc}_{\mathrm{gap}}
\right).
\end{equation}
Lower values indicate smaller group disparity. When reporting worst-group
performance, we use
\begin{equation}
\mathrm{WGA}
=
\min_{g\in\mathcal{G}} \mathrm{Acc}(g),
\end{equation}
where higher values indicate better worst-group utility.

\paragraph{Text classification.}
For text classification benchmarks, we follow the text fairness literature and
report worst-group macro-F1 \citep{chalkidis2022fairlex}, worst-class TPR gap
\citep{romanov2019s,hemamou2022delivering}, and RMS class TPR gap
\citep{ravfogel2020null,ravfogel2022linear}. These metrics are used because
text benchmarks such as occupation and sentiment classification are commonly
evaluated through class-wise group disparities rather than only aggregate binary
fairness gaps.

Let $\mathcal{C}$ be the label set. 
For each group $g$ and class $c$, let
$\mathrm{F1}_{g,c}$ denote the class-$c$ F1 score computed over
all samples belonging to group $g$. The group macro-F1 is
\begin{equation}
\mathrm{MacroF1}(g)
=
\frac{1}{|\mathcal{C}|}
\sum_{c\in\mathcal{C}}
\mathrm{F1}_{g,c}.
\end{equation}
Worst-group macro-F1 is
\begin{equation}
\mathrm{WgF1}
=
\min_{g\in\mathcal{G}}
\mathrm{MacroF1}(g).
\end{equation}
Higher values indicate better worst-group text classification performance.

For class-wise group recall disparity, define
\begin{equation}
\mathrm{TPR}_{g,c}
=
\mathbb{P}(\hat{Y}=c \mid Y=c,S=g).
\end{equation}
The TPR gap for class $c$ is
\begin{equation}
\mathrm{TPRGap}_{c}
=
\max_{g\in\mathcal{G}}\mathrm{TPR}_{g,c}
-
\min_{g\in\mathcal{G}}\mathrm{TPR}_{g,c}.
\end{equation}
The worst-class TPR gap is
\begin{equation}
\mathrm{WCTG}
=
\max_{c\in\mathcal{C}}
\mathrm{TPRGap}_{c}.
\end{equation}
The RMS class TPR gap is
\begin{equation}
\mathrm{RCTG}
=
\sqrt{
\frac{1}{|\mathcal{C}|}
\sum_{c\in\mathcal{C}}
\left(\mathrm{TPRGap}_{c}\right)^2
}.
\end{equation}
Lower WCTG and RCTG indicate smaller class-wise group disparity.

\paragraph{Regression.}
For regression, we report mean absolute error (MAE) as the main utility metric
and MAE gap for group error disparity \citep{han2022systematic}. The overall
MAE is
\begin{equation}
\mathrm{MAE}
=
\mathbb{E}\left[|Y-\hat{Y}|\right].
\end{equation}
The group-specific MAE is
\begin{equation}
\mathrm{MAE}(g)
=
\mathbb{E}\left[|Y-\hat{Y}| \mid S=g\right].
\end{equation}
Worst-group MAE and MAE gap are defined as
\begin{align}
\mathrm{WG\text{-}MAE}
&=
\max_{g\in\mathcal{G}}
\mathrm{MAE}(g), \\
\mathrm{MAE}_{\mathrm{gap}}
&=
\max_{g\in\mathcal{G}}
\mathrm{MAE}(g)
-
\min_{g\in\mathcal{G}}
\mathrm{MAE}(g).
\end{align}
Lower values indicate smaller prediction error and smaller group error disparity.
We also report $R^2$ as a regression utility metric:
\begin{equation}
R^2
=
1-
\frac{
\sum_i (y_i-\hat{y}_i)^2
}{
\sum_i (y_i-\bar{y})^2
}.
\end{equation}

\paragraph{Segmentation.}
For segmentation, we report Dice and HD95 as utility metrics. Dice measures
region overlap between the predicted and ground-truth masks, while HD95 measures
boundary error using the 95th-percentile Hausdorff distance. We additionally
report worst-group Dice for overlap fairness \citep{perez2025exploring} and
HD95 gap for boundary fairness \citep{vcevora2024quantifying}.

Let $P$ denote the predicted segmentation mask and $T$ the ground-truth mask.
The Dice score is
\begin{equation}
\mathrm{Dice}(P,T)
=
\frac{2|P\cap T|}{|P|+|T|}.
\end{equation}
For each group $g$, let $\mathrm{Dice}(g)$ denote the average Dice score over
samples with $S=g$. Worst-group Dice is
\begin{equation}
\mathrm{WG\text{-}Dice}
=
\min_{g\in\mathcal{G}}
\mathrm{Dice}(g).
\end{equation}
Higher values indicate better worst-group overlap quality.

For HD95, let $\partial P$ and $\partial T$ denote the boundaries of the
predicted and ground-truth masks. For point sets $A$ and $B$, define the directed
95th-percentile boundary distance as
\begin{equation}
h_{95}(A,B)
=
Q_{0.95}
\left(
\left\{
\min_{b\in B}\|a-b\|_2
:
a\in A
\right\}
\right),
\end{equation}
where $Q_{0.95}$ denotes the 95th percentile. The symmetric HD95 is
\begin{equation}
\mathrm{HD95}(P,T)
=
\max
\left\{
h_{95}(\partial P,\partial T),
h_{95}(\partial T,\partial P)
\right\}.
\end{equation}
For each group $g$, let $\mathrm{HD95}(g)$ denote the average HD95 over samples
with $S=g$. The HD95 gap is
\begin{equation}
\mathrm{HD95}_{\mathrm{gap}}
=
\max_{g\in\mathcal{G}}
\mathrm{HD95}(g)
-
\min_{g\in\mathcal{G}}
\mathrm{HD95}(g).
\end{equation}
Lower HD95 and lower HD95 gap indicate better boundary quality and smaller
boundary disparity.

%%%%%%%%%%%%%%%%%%%%%%%%%%%%%%%%%%%%%%%%%%%%%%%%%%%%%%%%%%%%%%%%%%
%% Additional Results
%%%%%%%%%%%%%%%%%%%%%%%%%%%%%%%%%%%%%%%%%%%%%%%%%%%%%%%%%%%%%%%%%%

\section{Additional Results}
\label{sect:appendix_results}

\subsection{Prototype-order ablation}
\label{app:prototype_order_ablation}

This ablation studies how the order of the prototype family affects feature-level sensitive-attribute removal on MOJI. The raw RFF representation is used as the reference, and higher-order \methodname variants add residual nuisance directions beyond the lower-order prototype subspace. As shown in Table~\ref{tab:moji_kinder_order_feature_ablation}, first-order prototypes induce only a small projection shift but still remove mean-level variation. Second- and third-order prototypes remove substantially more dependence and reduce HSIC along with covariance and third-moment gaps, at the cost of larger representation shifts and higher removed rank. This supports the use of prototype order as a controllable erasure-strength parameter.

\begin{table}[t]
\centering
\caption{
\textbf{Feature-level \methodname order ablation on MOJI using BERT.}
The RFF representation $\Phi(z)$ is used as the pre-projection reference. Higher prototype orders add residual nuisance directions beyond the lower-order subspace. Projection shift is $\|R-\Phi(z)\|_F/\|\Phi(z)\|_F$.
}
\small
\setlength{\tabcolsep}{4pt}
\scalebox{0.8}{
\begin{tabular}{lcccccc}
\toprule
Representation &
Projection Shift &
RBF-HSIC($R$, $S$) $\downarrow$ &
Mean Gap $\downarrow$ &
Cov. Gap $\downarrow$ &
3rd-Moment Gap $\downarrow$ &
Removed Rank \\
\midrule
Raw encoder $z$      & --    & --    & 1.164 & 1.457 & 1.793 & 0 \\
Raw RFF $\Phi(z)$    & 0.000 & 1.000 & 1.000 & 1.000 & 1.000 & 0 \\
\methodname-order 1  & 0.155 & 0.975 & 0.998 & 0.980 & 0.931 & 1 \\
\methodname-order 2  & 0.362 & 0.687 & 0.679 & 0.279 & 0.263 & 3 \\
\methodname-order 3  & 0.540 & 0.433 & 0.563 & 0.124 & 0.065 & 4 \\
\bottomrule
\end{tabular}
}
\label{tab:moji_kinder_order_feature_ablation}
\end{table}

\subsection{Feature-level bootstrap significance tests}
\label{app:feature_level_bootstrap}

To assess whether the feature-level differences in Fig.~\ref{fig:teaser}(c--d)
are significant, we performed paired bootstrap tests for \methodname vs the reported baselines on Moji dataset in Table~\ref{tab:moji_kinder_bootstrap_significance_compact}.
For each bootstrap sample, we recompute the normalized RBF-HSIC and average moment-gap
metrics. The improvement is defined as the baseline value minus the KINDER value, so
positive values favor KINDER. We report one-sided bootstrap $p$-values for positive
improvement and Benjamini--Hochberg corrected $q$-values across the tested comparisons. We observe that \methodname's shown improvements in Fig.~\ref{fig:teaser}(c--d) are mostly significant, with the exception of \methodname vs KCE average gaps where despite the 95\% CI being largely positive, it still holds zero inside it.

\begin{table}[t]
\centering
\caption{Paired bootstrap tests for the two feature-level summary metrics used in Fig.~\ref{fig:teaser}(c--d). Lower values are better. Improvement is baseline minus KINDER, so positive values favor KINDER.}
\small
\setlength{\tabcolsep}{4pt}
\resizebox{\columnwidth}{!}{%
\begin{tabular}{llrrrllll}
\toprule
Baseline & Metric & Baseline value & KINDER value & Improvement & 95\% CI & $p$ one-sided & $q$ BH & Signif. \\
\midrule
SUP & RBF-HSIC & 0.89 & 0.46 & 0.43 & [0.33, 0.70] & 0.003 & 0.005 & ** \\
SUP & Average gap & 0.96 & 0.27 & 0.69 & [0.64, 0.80] & 0.003 & 0.005 & ** \\
INLP & RBF-HSIC & 0.56 & 0.46 & 0.09 & [0.05, 0.46] & 0.010 & 0.013 & * \\
INLP & Average gap & 0.89 & 0.27 & 0.62 & [0.56, 0.75] & 0.003 & 0.005 & ** \\
RLACE & RBF-HSIC & 1.01 & 0.46 & 0.55 & [0.42, 0.77] & 0.003 & 0.005 & ** \\
RLACE & Average gap & 0.96 & 0.27 & 0.69 & [0.63, 0.79] & 0.003 & 0.005 & ** \\
KCE & RBF-HSIC & 1.10 & 0.46 & 0.64 & [0.50, 0.85] & 0.003 & 0.005 & ** \\
KCE & Average gap & 0.30 & 0.27 & 0.03 & [-0.03, 0.13] & 0.113 & 0.123 &  \\
\bottomrule
\end{tabular}
}
\label{tab:moji_kinder_bootstrap_significance_compact}
\end{table}

\subsection{Sequential MIMIC-IV classification ablation}
\label{app:mimic_seq_ablation}

Table~\ref{tab:mimiciv_seq_ablation} reports a sequential single-attribute removal ablation for MIMIC-IV classification. In the main experiment, Table~\ref{tab:mimiciv_group_intersectional}, \methodname removes the joint protected-attribute structure directly. Here, we instead apply each erasure method repeatedly, removing one protected attribute at a time in sequence, and then evaluate the final representation using the same marginal and intersectional fairness metrics. This tests whether methods originally designed around a single protected attribute remain stable when extended to multiple protected attributes through repeated erasure. The results show that repeated post-hoc removal can reduce some marginal gaps, but often introduces large utility loss or unstable intersectional behavior. In contrast, \methodname-seq (NH) preserves the baseline accuracy, gives the best F1 and worst-group accuracy, and obtains the lowest intersectional gap. The \methodname-seq vs \methodname-seq (NH) variants further show the head-mismatch effect. Replacing the native task head with a linear probe improves some gap values but substantially reduces utility.

\begin{table*}[t]
\centering
\scriptsize
\caption{\textbf{Sequential post-hoc MIMIC-IV classification ablation} using ResNet. Higher is better for accuracy, F1, and worst-group accuracy; lower is better for gap measures.
Best values in \textbf{bold} and second-best values are \underline{underlined}. 
}
\resizebox{\textwidth}{!}{
\begin{tabular}{lrrrrrrr}
\toprule
\textbf{Method}
& \textbf{Acc $\uparrow$}
& \textbf{F1 $\uparrow$}
& \textbf{Gender Gap $\downarrow$}
& \textbf{Race Gap $\downarrow$}
& \textbf{Age Gap $\downarrow$}
& \textbf{Inter. Gap $\downarrow$}
& \textbf{WorstGroupAcc $\uparrow$} \\
\midrule
ResNet      & \textbf{0.92} & 0.40 & 0.02 & 0.23 & \textbf{0.03} & 0.47 & \underline{0.66} \\
\midrule
+ INLP-seq      & 0.78 & 0.43 & 0.02 & 0.22 & 0.06 & 0.53 & 0.57 \\
+ RLACE-seq     & 0.78 & 0.43 & 0.03 & 0.23 & 0.07 & 0.53 & 0.58 \\
+ LEACE-seq     & \underline{0.81} & \underline{0.44} & \textbf{0.01} & 0.25 & 0.05 & 0.73 & 0.38 \\
+ KCE-seq       & 0.80 & 0.20 & 0.02 & \underline{0.19} & \underline{0.04} & 0.76 & 0.00 \\
\hline
+ \methodname-seq & 0.79 & 0.43 & 0.02 & \textbf{0.16 }& 0.08 & \underline{0.46} & 0.50 \\
+ \methodname-seq (NH)     & \textbf{0.92} & \textbf{0.46} & \textbf{0.01} & 0.31 & \textbf{0.03} & \textbf{0.44} & \textbf{0.69} \\
\bottomrule
\end{tabular}
}
\label{tab:mimiciv_seq_ablation}
\end{table*}

\subsection{Communities and Crime Dataset}
\label{sect:appendix_communities_and_crime_dataset}

Table~\ref{tab:communities_crime_mlp_plr_race_summary} reports the Communities and Crime regression results using MLP-PLR. This dataset tests whether concept erasure remains useful in a low-dimensional tabular regression setting where protected-group structure is tied to community-level socioeconomic variables. \methodname achieves the best overall utility, with the lowest MAE and highest \(R^2\), and also gives the best race worst-group error. SUP and RLACE obtain slightly lower race-gap values, but \methodname remains close while preserving stronger predictive performance. This result is consistent with the main regression findings. Aggressive post-hoc removal can reduce some fairness gaps, but often loses substantial task information, while \methodname gives a stronger utility-preserving tradeoff.

\begin{table}[t]
\centering
\caption{\textbf{Communities and Crime regression results using MLP-PLR.} Lower is better for MAE, worst-group MAE, and MAE gap; higher is better for $R^2$. Best values in \textbf{bold} and second-best values are \underline{underlined}. 
\label{tab:communities_crime_mlp_plr_race_summary}
}
\small
%\resizebox{\textwidth}{!}{
\begin{tabular}{lrrrr}
\toprule
\textbf{Method}
& \textbf{MAE $\downarrow$}
& \textbf{$R^2$ (\%) $\uparrow$}
& \textbf{Race WG $\downarrow$}
& \textbf{Race Gap $\downarrow$} \\
\midrule
MLP-PLR & \underline{0.093} & \underline{63.8} & 0.154 & 0.090 \\
+ INLP  & 0.148 & 18.2 & 0.288 & 0.164 \\
+ LEACE & 0.157 & 11.7 & 0.359 & 0.225 \\
+ RLACE & 0.094 & 63.8 & 0.152 & \underline{0.071} \\
+ KCE   & 0.096 & 60.2 & 0.190 & 0.113 \\
+ SUP   & 0.094 & 63.8 & \underline{0.149} & \textbf{0.070} \\
\cmidrule(lr){1-5}
+ \methodname & \textbf{0.092} & \textbf{65.3} & \textbf{0.145} & 0.079 \\
\bottomrule
\end{tabular}
%}
\end{table}

\subsection{Three-seed downstream results}
\label{app:rebuttal_three_seed}

The following tables report three-seed downstream results. These results are reported separately from the main tables because the revised evaluation selected configurations using validation performance averaged across three seeds. Significance marks indicate one-sided paired $t$-tests, with $^{*}p<0.10$, $^{**}p<0.05$, and $^{***}p<0.01$.

\begin{table*}[t]
\centering
\setlength{\belowcaptionskip}{4pt}
\caption{\textbf{MIMIC-IV classification} three-seed results using TabM. Acc and F1 are performance metrics and the remaining columns are fairness metrics. Arrows indicate the preferred direction. Entries show the reported mean $\pm$ standard deviation. Best values are in \textbf{bold}, second-best values are \underline{underlined}.}
\label{tab:rebuttal_mimic_classification}
\scriptsize
\resizebox{\textwidth}{!}{%
\begin{tabular}{lccccccc}
\toprule
\textbf{Method}
& \textbf{Acc $\uparrow$}
& \textbf{F1 $\uparrow$}
& \textbf{Gender Gap $\downarrow$}
& \textbf{Race Gap $\downarrow$}
& \textbf{Age Gap $\downarrow$}
& \textbf{Inter. Gap $\downarrow$}
& \textbf{WG Inter. $\uparrow$} \\
\midrule
TabM & \underline{91.95} $\pm$ 0.07 & 41.72 $\pm$ 1.49 & 1.15 $\pm$ 0.33 & 20.63 $\pm$ 1.82 & \underline{3.25} $\pm$ 1.01 & 70.83 $\pm$ 0.00 & 50.00 $\pm$ 0.00 \\
+ INLP & 77.67 $\pm$ 3.59 & 41.95 $\pm$ 2.96 & 1.70 $\pm$ 0.79 & 21.48 $\pm$ 2.31 & 5.62 $\pm$ 0.91 & 49.63 $\pm$ 3.97 & \underline{57.82} $\pm$ 10.80 \\
+ RLACE & 77.65 $\pm$ 2.93 & 42.14 $\pm$ 2.18 & 1.73 $\pm$ 0.52 & 22.38 $\pm$ 3.04 & 7.95 $\pm$ 0.91 & 47.98 $\pm$ 2.04 & 57.64 $\pm$ 8.87 \\
+ LEACE & 77.82 $\pm$ 2.52 & 42.20 $\pm$ 1.67 & 1.98 $\pm$ 0.39 & 20.88 $\pm$ 1.47 & 6.77 $\pm$ 1.22 & \textbf{44.72} $\pm$ 2.68 & 56.06 $\pm$ 8.37 \\
+ KCE & 78.83 $\pm$ 2.04 & 17.46 $\pm$ 1.18 & 1.26 $\pm$ 0.52 & \textbf{16.96} $\pm$ 6.68 & 4.34 $\pm$ 0.35 & 74.42 $\pm$ 4.04 & 46.67 $\pm$ 7.45 \\
+ SAL & 77.01 $\pm$ 2.34 & 41.66 $\pm$ 1.61 & 1.85 $\pm$ 0.60 & 21.63 $\pm$ 3.08 & 7.02 $\pm$ 0.69 & 47.41 $\pm$ 3.49 & 51.82 $\pm$ 6.89 \\
\hline
\methodname post-hoc & 78.81 $\pm$ 1.05 & 43.01 $\pm$ 1.00 & 1.37 $\pm$ 0.33 & 20.63 $\pm$ 3.06 & 6.82 $\pm$ 0.98 & 46.69 $\pm$ 4.61 & 57.58 $\pm$ 13.12 \\
+ \methodname & 79.82 $\pm$ 1.49 & \underline{44.48} $\pm$ 1.27\textsuperscript{*} & \underline{1.14} $\pm$ 0.19\textsuperscript{*} & \underline{20.51} $\pm$ 2.74\textsuperscript{**} & 7.17 $\pm$ 0.95 & \underline{46.59} $\pm$ 0.00 & \textbf{63.64} $\pm$ 0.00 \\
+ \methodname (NH) & \textbf{92.05} $\pm$ 0.07\textsuperscript{***} & \textbf{46.17} $\pm$ 1.35\textsuperscript{*} & \textbf{0.88} $\pm$ 0.39\textsuperscript{*} & 22.10 $\pm$ 8.28 & \textbf{2.23} $\pm$ 0.27\textsuperscript{***} & 72.22 $\pm$ 2.41 & 50.00 $\pm$ 0.00 \\
\bottomrule
\end{tabular}
}
\end{table*}

\begin{table}[t]
\centering
\caption{\textbf{MIMIC-IV regression} three-seed results using MLP-PLR. MAE and $R^2$ measure performance; worst-group MAE columns measure fairness. Arrows indicate the preferred direction. Entries show the reported mean $\pm$ standard deviation. Best values are in \textbf{bold}, second-best values are \underline{underlined}.}
\label{tab:rebuttal_mimic_regression}
\small
\resizebox{\columnwidth}{!}{%
\begin{tabular}{lcccccc}
\toprule
\textbf{Method}
& \textbf{MAE $\downarrow$}
& \textbf{$R^2$ $\uparrow$}
& \textbf{Race WG-MAE $\downarrow$}
& \textbf{Gender WG-MAE $\downarrow$}
& \textbf{Age WG-MAE $\downarrow$}
& \textbf{Inter. WG-MAE $\downarrow$} \\
\midrule
MLP-PLR & \underline{2.62} $\pm$ 0.05 & \textbf{0.12} $\pm$ 0.00 & \underline{3.39} $\pm$ 0.06 & \underline{2.64} $\pm$ 0.05 & 2.88 $\pm$ 0.06 & \underline{8.00} $\pm$ 1.21 \\
+ INLP & 2.73 $\pm$ 0.13 & 0.09 $\pm$ 0.05 & 3.50 $\pm$ 0.13 & 2.76 $\pm$ 0.14 & 2.98 $\pm$ 0.16 & 8.29 $\pm$ 1.87 \\
+ LEACE & 2.70 $\pm$ 0.01 & 0.08 $\pm$ 0.02 & 3.48 $\pm$ 0.06 & 2.74 $\pm$ 0.01 & \underline{2.84} $\pm$ 0.06 & 8.49 $\pm$ 2.16 \\
+ SUP & 2.68 $\pm$ 0.00 & 0.08 $\pm$ 0.03 & 3.51 $\pm$ 0.05 & 2.71 $\pm$ 0.02 & 2.95 $\pm$ 0.08 & \textbf{7.40} $\pm$ 1.96 \\
+ SAL & 2.70 $\pm$ 0.01 & 0.08 $\pm$ 0.03 & 3.54 $\pm$ 0.07 & 2.71 $\pm$ 0.02 & 2.92 $\pm$ 0.09 & 8.27 $\pm$ 2.14 \\
\midrule
+ \methodname & \textbf{2.54} $\pm$ 0.04\textsuperscript{**} & \underline{0.11} $\pm$ 0.00\textsuperscript{*} & \textbf{3.31} $\pm$ 0.06\textsuperscript{***} & \textbf{2.55} $\pm$ 0.05\textsuperscript{**} & \textbf{2.78} $\pm$ 0.07\textsuperscript{**} & 8.80 $\pm$ 0.16 \\
\bottomrule
\end{tabular}
}
\end{table}

\begin{table*}[t]
\centering
\caption{\textbf{FairSeg segmentation} three-seed results. Dice and HD95 measure performance; D-WG denotes worst-group Dice and Dice Gap denotes subgroup Dice disparity. Arrows indicate the preferred direction. Entries show the reported mean $\pm$ standard deviation. Best values are in \textbf{bold}, second-best values are \underline{underlined}.}
\label{tab:rebuttal_fairseg}
\small
\resizebox{\textwidth}{!}{%
\begin{tabular}{lcc|cc|cc|cc|cc}
\toprule
\multirow{2}{*}{\textbf{Method}}
& \multicolumn{2}{c|}{\textbf{Perf.}}
& \multicolumn{2}{c|}{\textbf{Race}}
& \multicolumn{2}{c|}{\textbf{Gender}}
& \multicolumn{2}{c|}{\textbf{Ethnicity}}
& \multicolumn{2}{c}{\textbf{Inter.}} \\
\cmidrule(lr){2-3}
\cmidrule(lr){4-5}
\cmidrule(lr){6-7}
\cmidrule(lr){8-9}
\cmidrule(lr){10-11}
& \textbf{Dice $\uparrow$}
& \textbf{HD95 $\downarrow$}
& \textbf{D-WG $\uparrow$}
& \textbf{Dice Gap $\downarrow$}
& \textbf{D-WG $\uparrow$}
& \textbf{Dice Gap $\downarrow$}
& \textbf{D-WG $\uparrow$}
& \textbf{Dice Gap $\downarrow$}
& \textbf{D-WG $\uparrow$}
& \textbf{Dice Gap $\downarrow$} \\
\midrule
Backbone & 83.12 $\pm$ 1.22 & 5.12 $\pm$ 1.19 & 78.25 $\pm$ 0.97 & 4.03 $\pm$ 0.29 & 82.93 $\pm$ 1.17 & 0.37 $\pm$ 0.24 & 82.02 $\pm$ 0.53 & 1.60 $\pm$ 0.34 & 65.88 $\pm$ 2.62 & 16.00 $\pm$ 10.18 \\
+ INLP & 83.42 $\pm$ 1.07 & 4.82 $\pm$ 0.72 & 78.43 $\pm$ 1.45 & 3.85 $\pm$ 0.66 & 83.22 $\pm$ 0.91 & 0.50 $\pm$ 0.35 & 82.38 $\pm$ 0.87 & \textbf{1.32} $\pm$ 0.82 & 66.53 $\pm$ 4.33 & 13.35 $\pm$ 5.81 \\
+ RLACE & 83.70 $\pm$ 0.72 & 4.72 $\pm$ 0.64 & 78.97 $\pm$ 1.01 & 3.80 $\pm$ 0.67 & 83.58 $\pm$ 0.69 & \textbf{0.20} $\pm$ 0.12 & 82.52 $\pm$ 0.96 & 1.42 $\pm$ 0.86 & 67.92 $\pm$ 3.83 & 11.50 $\pm$ 3.99 \\
+ LEACE & 83.68 $\pm$ 0.81 & 4.60 $\pm$ 0.73 & 79.10 $\pm$ 1.28 & \textbf{3.18} $\pm$ 1.32 & 83.52 $\pm$ 0.81 & 0.32 $\pm$ 0.19 & 82.52 $\pm$ 1.11 & 1.58 $\pm$ 0.56 & \underline{69.22} $\pm$ 5.34 & \underline{8.22} $\pm$ 2.42 \\
+ SUP & 83.60 $\pm$ 0.82 & 4.90 $\pm$ 0.88 & 78.68 $\pm$ 1.12 & 3.97 $\pm$ 0.34 & 83.50 $\pm$ 0.70 & 0.30 $\pm$ 0.36 & 82.55 $\pm$ 0.84 & 1.60 $\pm$ 0.22 & 65.65 $\pm$ 2.86 & 16.98 $\pm$ 9.37 \\
+ KCE & \textbf{84.05} $\pm$ 0.37 & 4.53 $\pm$ 0.05 & \underline{79.25} $\pm$ 0.66 & 3.50 $\pm$ 1.07 & \textbf{83.88} $\pm$ 0.39 & \underline{0.25} $\pm$ 0.10 & \textbf{83.00} $\pm$ 0.27 & \underline{1.38} $\pm$ 0.59 & 67.58 $\pm$ 5.54 & 10.95 $\pm$ 3.27 \\
+ SAL & 83.97 $\pm$ 0.68 & \underline{4.43} $\pm$ 0.59 & 78.63 $\pm$ 0.86 & 4.03 $\pm$ 0.06 & 83.63 $\pm$ 0.50 & 0.40 $\pm$ 0.17 & 82.27 $\pm$ 0.78 & 2.07 $\pm$ 0.15 & 65.87 $\pm$ 0.61 & 9.57 $\pm$ 1.55 \\
\cmidrule(lr){1-11}
+ \methodname & \underline{84.00} $\pm$ 0.22 & \textbf{4.18} $\pm$ 0.29\textsuperscript{*} & \textbf{79.40} $\pm$ 0.71\textsuperscript{*} & \underline{3.38} $\pm$ 1.23 & \underline{83.78} $\pm$ 0.15 & \underline{0.25} $\pm$ 0.13 & \underline{82.80} $\pm$ 0.62 & 1.45 $\pm$ 0.91 & \textbf{70.18} $\pm$ 5.38\textsuperscript{***} & \textbf{7.43} $\pm$ 1.52\textsuperscript{**} \\
\bottomrule
\end{tabular}
}
\end{table*}

\begin{table}[t]
\centering
\caption{\textbf{BIOS classification} three-seed results using BERT. WgF1 denotes worst-group macro-F1, WCTG denotes worst-class TPR gap, and RCTG denotes RMS class TPR gap. Arrows indicate the preferred direction. Entries show the reported mean $\pm$ standard deviation. Best values are in \textbf{bold}, second-best values are \underline{underlined}.}
\label{tab:rebuttal_bios}
\scriptsize
\setlength{\tabcolsep}{1.8pt}
\resizebox{0.73\columnwidth}{!}{%
\begin{tabular}{@{}lccccc@{}}
\toprule
\textbf{Method}
& \textbf{Acc $\uparrow$}
& \textbf{F1 $\uparrow$}
& \textbf{WgF1 $\uparrow$}
& \textbf{WCTG $\downarrow$}
& \textbf{RCTG $\downarrow$} \\
\midrule
BERT & \underline{80.12} $\pm$ 0.04 & \underline{70.98} $\pm$ 0.05 & 66.42 $\pm$ 0.07 & 52.40 $\pm$ 0.18 & 19.58 $\pm$ 0.11 \\
+ INLP & 79.04 $\pm$ 0.05 & 68.53 $\pm$ 0.04 & 66.91 $\pm$ 0.11 & \textbf{46.64} $\pm$ 0.22 & 14.65 $\pm$ 0.21 \\
+ RLACE & 79.82 $\pm$ 0.06 & 70.52 $\pm$ 0.11 & \underline{67.70} $\pm$ 0.01 & 51.28 $\pm$ 0.65 & 16.22 $\pm$ 0.07 \\
+ LEACE & 79.38 $\pm$ 0.04 & 70.08 $\pm$ 0.09 & 67.05 $\pm$ 0.05 & 47.90 $\pm$ 0.15 & \textbf{12.96} $\pm$ 0.07 \\
+ KCE & 78.22 $\pm$ 2.25 & 70.40 $\pm$ 2.89 & 66.63 $\pm$ 3.50 & 49.78 $\pm$ 2.97 & 16.25 $\pm$ 3.10 \\
+ SUP & 79.52 $\pm$ 0.08 & 70.18 $\pm$ 0.14 & 65.82 $\pm$ 0.18 & 51.35 $\pm$ 0.45 & 18.31 $\pm$ 0.20 \\
+ SAL & 79.11 $\pm$ 0.01 & 69.95 $\pm$ 0.08 & 66.95 $\pm$ 0.13 & \underline{47.81} $\pm$ 0.30 & \underline{13.05} $\pm$ 0.20 \\
\cmidrule(lr){1-6}
+ \methodname & \textbf{82.08} $\pm$ 0.07\textsuperscript{***} & \textbf{72.72} $\pm$ 0.20\textsuperscript{***} & \textbf{71.65} $\pm$ 0.32\textsuperscript{***} & 47.98 $\pm$ 2.32\textsuperscript{*} & 14.28 $\pm$ 0.69\textsuperscript{***} \\
\bottomrule
\end{tabular}
}
\end{table}

\subsection{Matched end-to-end fairness interventions}
\label{app:rebuttal_matched_end_to_end}

We additionally compare \methodname with end-to-end fairness interventions using the same TabM backbone, native task head, and matched random seeds. These methods are reported separately from the post-hoc concept-erasure baselines.

\begin{table*}[t]
\centering
\setlength{\belowcaptionskip}{4pt}
\caption{\textbf{MIMIC-IV classification} matched end-to-end results using TabM and the native task head. Acc and F1 are performance metrics and the remaining columns are fairness metrics. Arrows indicate the preferred direction. Entries show the reported three-seed mean $\pm$ standard deviation. Best values are in \textbf{bold}, second-best values are \underline{underlined}.}
\label{tab:rebuttal_matched_end_to_end}
\scriptsize
\resizebox{\textwidth}{!}{%
\begin{tabular}{lccccccc}
\toprule
\textbf{Method}
& \textbf{Acc $\uparrow$}
& \textbf{F1 $\uparrow$}
& \textbf{Gender Gap $\downarrow$}
& \textbf{Race Gap $\downarrow$}
& \textbf{Age Gap $\downarrow$}
& \textbf{Inter. Gap $\downarrow$}
& \textbf{WG Inter. $\uparrow$} \\
\midrule
TabM & \underline{91.95} $\pm$ 0.07 & 41.72 $\pm$ 1.49 & 1.15 $\pm$ 0.33 & \textbf{20.63} $\pm$ 1.82 & 3.25 $\pm$ 1.01 & \underline{70.83} $\pm$ 0.00 & 50.00 $\pm$ 0.00 \\
+ FR-Train & 91.67 $\pm$ 0.05 & 38.39 $\pm$ 3.35 & \textbf{0.34} $\pm$ 0.20 & \underline{21.61} $\pm$ 3.31 & 3.45 $\pm$ 0.99 & \underline{70.83} $\pm$ 0.00 & 50.00 $\pm$ 0.00 \\
+ FairMixup & 80.59 $\pm$ 0.59 & \underline{44.66} $\pm$ 0.42 & 1.34 $\pm$ 0.62 & 24.48 $\pm$ 1.34 & 7.38 $\pm$ 0.51 & \textbf{50.02} $\pm$ 2.66 & \textbf{60.91} $\pm$ 6.10 \\
+ GroupDRO & 91.31 $\pm$ 0.27 & 40.58 $\pm$ 3.73 & 1.07 $\pm$ 0.55 & 26.36 $\pm$ 11.44 & \underline{3.11} $\pm$ 1.06 & 71.00 $\pm$ 2.66 & \underline{56.00} $\pm$ 13.42 \\
+ FairBatch & 91.65 $\pm$ 0.13 & 39.66 $\pm$ 3.22 & \underline{0.47} $\pm$ 0.33 & 29.96 $\pm$ 6.90 & 4.11 $\pm$ 1.30 & 73.33 $\pm$ 2.28 & 50.00 $\pm$ 0.00 \\
\hline
+ \methodname (NH) & \textbf{92.05} $\pm$ 0.07\textsuperscript{**} & \textbf{46.17} $\pm$ 1.35\textsuperscript{*} & 0.88 $\pm$ 0.39 & 22.10 $\pm$ 8.28 & \textbf{2.23} $\pm$ 0.27\textsuperscript{*} & 72.22 $\pm$ 2.41 & 50.00 $\pm$ 0.00 \\
\bottomrule
\end{tabular}
}
\end{table*}

\subsection{Multi-seed results and SAL comparison}
\label{app:rebuttal_first_order_sal}

We additionally compare \methodname directly with SAL across the four settings. SAL~\cite{shao2023gold} is a kernelized spectral projection baseline for attribute removal, and it has partial methodological overlap limited to a special case of first-order \methodname. For categorical protected attributes, SAL's cross-covariance directions align with prevalence-weighted group-mean deviations under matched feature maps, encodings, centering, weighting, and retained rank. The three-seed results show improvements in the main utility metrics across all four settings and several fairness metrics. For MIMIC-IV classification, we distinguish the post-hoc and native-head (NH) configurations.

\begin{table*}[t]
\centering
\setlength{\belowcaptionskip}{4pt}
\caption{\textbf{MIMIC-IV classification} first-order \methodname and SAL comparison. Entries show the reported three-seed mean $\pm$ standard deviation. Best values are in \textbf{bold}, second-best values are \underline{underlined}.}
\label{tab:rebuttal_first_mimic_class}
\scriptsize
\resizebox{\textwidth}{!}{%
\begin{tabular}{lccccccc}
\toprule
\textbf{Method}
& \textbf{Acc $\uparrow$}
& \textbf{F1 $\uparrow$}
& \textbf{Gender Gap $\downarrow$}
& \textbf{Race Gap $\downarrow$}
& \textbf{Age Gap $\downarrow$}
& \textbf{Inter. Gap $\downarrow$}
& \textbf{WG Inter. $\uparrow$} \\
\midrule
+ SAL & 77.01 $\pm$ 2.34 & 41.66 $\pm$ 1.61 & 1.85 $\pm$ 0.60 & \underline{21.63} $\pm$ 3.08 & 7.02 $\pm$ 0.69 & \textbf{47.41} $\pm$ 3.49 & 51.82 $\pm$ 6.89 \\
\hline
+ \methodname post-hoc (Order 1) & 80.22 $\pm$ 1.07 & 44.32 $\pm$ 0.93 & \underline{1.22} $\pm$ 0.68 & \textbf{21.15} $\pm$ 2.57 & 6.97 $\pm$ 0.65 & 56.53 $\pm$ 13.53 & \underline{55.56} $\pm$ 9.62 \\
+ \methodname (Order 1) & \underline{80.72} $\pm$ 1.52 & \underline{45.38} $\pm$ 1.16 & 1.51 $\pm$ 0.23 & 22.61 $\pm$ 3.08 & \underline{6.24} $\pm$ 0.52 & \underline{49.81} $\pm$ 3.13 & \textbf{64.65} $\pm$ 1.75 \\
+ \methodname (NH, Order 1) & \textbf{91.93} $\pm$ 0.02 & \textbf{48.62} $\pm$ 1.50 & \textbf{1.17} $\pm$ 0.33 & 28.08 $\pm$ 3.75 & \textbf{2.87} $\pm$ 0.42 & 75.00 $\pm$ 0.00 & 50.00 $\pm$ 0.00 \\
\bottomrule
\end{tabular}
}
\end{table*}

\begin{table}[t]
\centering
\caption{\textbf{MIMIC-IV regression} first-order \methodname and SAL comparison using MLP-PLR. Entries show the reported three-seed mean $\pm$ standard deviation. Best values are in \textbf{bold}, second-best values are \underline{underlined}.}
\label{tab:rebuttal_first_mimic_reg}
\small
\resizebox{\columnwidth}{!}{%
\begin{tabular}{lcccccc}
\toprule
\textbf{Method}
& \textbf{MAE $\downarrow$}
& \textbf{$R^2$ $\uparrow$}
& \textbf{Race WG-MAE $\downarrow$}
& \textbf{Gender WG-MAE $\downarrow$}
& \textbf{Age WG-MAE $\downarrow$}
& \textbf{Inter. WG-MAE $\downarrow$} \\
\midrule
+ SAL & \underline{2.70} $\pm$ 0.01 & \underline{0.08} $\pm$ 0.03 & \underline{3.54} $\pm$ 0.07 & \underline{2.71} $\pm$ 0.02 & \underline{2.92} $\pm$ 0.09 & \textbf{8.27} $\pm$ 2.14 \\
\midrule
+ \methodname (Order 1) & \textbf{2.56} $\pm$ 0.09 & \textbf{0.12} $\pm$ 0.00 & \textbf{3.37} $\pm$ 0.16 & \textbf{2.58} $\pm$ 0.10 & \textbf{2.81} $\pm$ 0.10 & \underline{10.12} $\pm$ 0.25 \\
\bottomrule
\end{tabular}
}
\end{table}

\begin{table}[t]
\centering
\caption{\textbf{BIOS classification} first-order \methodname and SAL comparison using BERT. Entries show the reported three-seed mean $\pm$ standard deviation. Best values are in \textbf{bold}, second-best values are \underline{underlined}.}
\label{tab:rebuttal_first_bios}
\scriptsize
\setlength{\tabcolsep}{1.8pt}
\resizebox{0.73\columnwidth}{!}{%
\begin{tabular}{@{}lccccc@{}}
\toprule
\textbf{Method}
& \textbf{Acc $\uparrow$}
& \textbf{F1 $\uparrow$}
& \textbf{WgF1 $\uparrow$}
& \textbf{WCTG $\downarrow$}
& \textbf{RCTG $\downarrow$} \\
\midrule
+ SAL & \underline{79.11} $\pm$ 0.01 & \underline{69.95} $\pm$ 0.08 & \underline{66.95} $\pm$ 0.13 & \underline{47.81} $\pm$ 0.30 & \textbf{13.05} $\pm$ 0.20 \\
\cmidrule(lr){1-6}
+ \methodname (Order 1) & \textbf{82.00} $\pm$ 0.04 & \textbf{73.54} $\pm$ 0.26 & \textbf{70.99} $\pm$ 0.31 & \textbf{47.79} $\pm$ 2.54 & \underline{13.98} $\pm$ 0.68 \\
\bottomrule
\end{tabular}
}
\end{table}

\begin{table*}[t]
\centering
\caption{\textbf{FairSeg segmentation} first-order \methodname and SAL comparison. D-WG denotes worst-group Dice and Dice Gap denotes subgroup Dice disparity. Entries show the reported three-seed mean $\pm$ standard deviation. Best values are in \textbf{bold}, second-best values are \underline{underlined}.}
\label{tab:rebuttal_first_fairseg}
\small
\resizebox{\textwidth}{!}{%
\begin{tabular}{lcc|cc|cc|cc|cc}
\toprule
\multirow{2}{*}{\textbf{Method}}
& \multicolumn{2}{c|}{\textbf{Perf.}}
& \multicolumn{2}{c|}{\textbf{Race}}
& \multicolumn{2}{c|}{\textbf{Gender}}
& \multicolumn{2}{c|}{\textbf{Ethnicity}}
& \multicolumn{2}{c}{\textbf{Inter.}} \\
\cmidrule(lr){2-3}
\cmidrule(lr){4-5}
\cmidrule(lr){6-7}
\cmidrule(lr){8-9}
\cmidrule(lr){10-11}
& \textbf{Dice $\uparrow$}
& \textbf{HD95 $\downarrow$}
& \textbf{D-WG $\uparrow$}
& \textbf{Dice Gap $\downarrow$}
& \textbf{D-WG $\uparrow$}
& \textbf{Dice Gap $\downarrow$}
& \textbf{D-WG $\uparrow$}
& \textbf{Dice Gap $\downarrow$}
& \textbf{D-WG $\uparrow$}
& \textbf{Dice Gap $\downarrow$} \\
\midrule
+ SAL & \underline{83.97} $\pm$ 0.68 & \underline{4.43} $\pm$ 0.59 & \underline{78.63} $\pm$ 0.86 & \underline{4.03} $\pm$ 0.06 & \underline{83.63} $\pm$ 0.50 & \underline{0.40} $\pm$ 0.17 & \underline{82.27} $\pm$ 0.78 & \underline{2.07} $\pm$ 0.15 & \underline{65.87} $\pm$ 0.61 & \underline{9.57} $\pm$ 1.55 \\
\cmidrule(lr){1-11}
+ \methodname (Order 1) & \textbf{84.00} $\pm$ 0.22 & \textbf{4.18} $\pm$ 0.29 & \textbf{79.40} $\pm$ 0.71 & \textbf{3.38} $\pm$ 1.23 & \textbf{83.78} $\pm$ 0.15 & \textbf{0.25} $\pm$ 0.13 & \textbf{82.80} $\pm$ 0.62 & \textbf{1.45} $\pm$ 0.91 & \textbf{70.18} $\pm$ 5.38 & \textbf{7.43} $\pm$ 1.52 \\
\bottomrule
\end{tabular}
}
\end{table*}

\clearpage

%%%%%%%%%%%%%%%%%%%%%%%%%%%%%%%%%%%%%%%%%%%%%%%%%%%%%%%%%%%%%%%%%%%%%%%%%%%

\section{Limitations}
\label{sect:appendix_limitation}

\methodname assumes access to protected-attribute labels during training or post-hoc prototype estimation, which is standard in fairness evaluation but may not hold in all deployment settings. Its effectiveness also depends on the chosen prototype functionals and RFF approximation. Richer prototypes can capture more group-conditioned structure, but may increase computation and remove more task-relevant variation if the protected attributes are strongly entangled with the target. In addition, while our theory gives prototype-space independence, the full probabilistic independence requires the much stronger condition that the prototype family is defined in a way that identifies the relevant group laws. Future work should explore additional prototype families, settings with missing or noisy sensitive attributes, and larger-scale evaluations under severe intersectional sparsity.

\section{Impact Statement}
\label{sect:appendix_impact_statement}
Our work recognizes that ML models can cause negative societal impacts if not addressed. 
Thus, we present a method, 
\methodname
to promote fairer machine learning (ML) methods that can be fully generalizable across a range of tasks, data types and training settings. 
By developing \methodname, we aim to reduce the ML model performance gaps across different demographic groups at both the group and intersectional level, and thus generating a positive societal impact. We acknowledge that the sensitive attributes may not capture all dimensions of identities and future work should take these into consideration.

\clearpage
\section*{NeurIPS Paper Checklist}

%%% BEGIN INSTRUCTIONS %%%

%%% END INSTRUCTIONS %%%

\begin{enumerate}

\item {\bf Claims}
    \item[] Question: Do the main claims made in the abstract and introduction accurately reflect the paper's contributions and scope?
    \item[] Answer:  \answerYes{}% Replace by \answerYes{}, \answerNo{}, or \answerNA{}.
    \item[] Justification: The abstract and introduction state four contributions.
    Our first key contribution is outlined in Section
     \ref{sect:method_projection},  
     our second contribution is outlined in  Section \ref{sect:method_prototype}, \ref{sect:method_projection}. 
     The claim is that \methodname is generalizable and works across diverse settings, tasks and modalities in Section \ref{sect:results}.


\item {\bf Limitations}
    \item[] Question: Does the paper discuss the limitations of the work performed by the authors?
    \item[] Answer: \answerYes{} % Replace by \answerYes{}, \answerNo{}, or \answerNA{}.
    \item[] Justification: Discussion and Conclusion Section (Section \ref{sect:discussion}) has a dedicated coverage of the limitations of our work which we further expand on in Appendix \ref{sect:appendix_limitation}.

\item {\bf Theory assumptions and proofs}
    \item[] Question: For each theoretical result, does the paper provide the full set of assumptions and a complete (and correct) proof?
    \item[] Answer: \answerYes{} % Replace by \answerYes{}, \answerNo{}, or \answerNA{}.
    \item[] Justification: The theoretical discussions in the paper list the assumptions made. %\justificationTODO{}

    \item {\bf Experimental result reproducibility}
    \item[] Question: Does the paper fully disclose all the information needed to reproduce the main experimental results of the paper to the extent that it affects the main claims and/or conclusions of the paper (regardless of whether the code and data are provided or not)?
    \item[] Answer: \answerYes{} % Replace by \answerYes{}, \answerNo{}, or \answerNA{}.
    \item[] Justification: All experimental details are included. Furthermore, reproduction codes and trained models are available at: https://github.com/abtinmU/KINDER
    %will be shared upon acceptance. %\justificationTODO{}

\item {\bf Open access to data and code}
    \item[] Question: Does the paper provide open access to the data and code, with sufficient instructions to faithfully reproduce the main experimental results, as described in supplemental material?
    \item[] Answer: \answerYes{}
    %\answerTODO{} % Replace by \answerYes{}, \answerNo{}, or \answerNA{}.
    \item[] Justification: All datasets used are publicly available. The code that can be used to reproduce our results is 
    shared at: https://github.com/abtinmU/KINDER
    %\justificationTODO{}

\item {\bf Experimental setting/details}
    \item[] Question: Does the paper specify all the training and test details (e.g., data splits, hyperparameters, how they were chosen, type of optimizer) necessary to understand the results?
    \item[] Answer: \answerYes{} % Replace by \answerYes{}, \answerNo{}, or \answerNA{}.
    \item[] Justification: All the training setup are included inside the appendix. 

\item {\bf Experiment statistical significance}
    \item[] Question: Does the paper report error bars suitably and correctly defined or other appropriate information about the statistical significance of the experiments?
    \item[] Answer: \answerYes{} % Replace by \answerYes{}, \answerNo{}, or \answerNA{}.
    \item[] Justification: Significant test results are included inside the appendix.

\item {\bf Experiments compute resources}
    \item[] Question: For each experiment, does the paper provide sufficient information on the computer resources (type of compute workers, memory, time of execution) needed to reproduce the experiments?
    \item[] Answer: \answerYes{} % Replace by \answerYes{}, \answerNo{}, or \answerNA{}.
    \item[] Justification: Details are included inside the appendix.
    
\item {\bf Code of ethics}
    \item[] Question: Does the research conducted in the paper conform, in every respect, with the NeurIPS Code of Ethics \url{https://neurips.cc/public/EthicsGuidelines}?
    \item[] Answer: \answerYes{} % Replace by \answerYes{}, \answerNo{}, or \answerNA{}.
    \item[] Justification: The research conducted in this paper conforms with the NeurIPS Code of Ethics.

\item {\bf Broader impacts}
    \item[] Question: Does the paper discuss both potential positive societal impacts and negative societal impacts of the work performed?
    \item[] Answer: \answerYes{}
    %\answerTODO{} % Replace by \answerYes{}, \answerNo{}, or \answerNA{}.
    \item[] Justification: A dedicated impact section has been included under Section \ref{sect:appendix_impact_statement}.
    %\justificationTODO{}
    
\item {\bf Safeguards}
    \item[] Question: Does the paper describe safeguards that have been put in place for responsible release of data or models that have a high risk for misuse (e.g., pre-trained language models, image generators, or scraped datasets)?
    \item[] Answer: \answerNA{} % Replace by \answerYes{}, \answerNo{}, or \answerNA{}.
    \item[] Justification: We do not release data or generative models. %\justificationTODO{}

\item {\bf Licenses for existing assets}
    \item[] Question: Are the creators or original owners of assets (e.g., code, data, models), used in the paper, properly credited and are the license and terms of use explicitly mentioned and properly respected?
    \item[] Answer: \answerYes{} % Replace by \answerYes{}, \answerNo{}, or \answerNA{}.
    \item[] Justification: All datasets used are publicly available for research purposes. All datasets and baseline models have been cited properly. %\justificationTODO{}

\item {\bf New assets}
    \item[] Question: Are new assets introduced in the paper well documented and is the documentation provided alongside the assets?
    \item[] Answer: \answerYes{} % Replace by \answerYes{}, \answerNo{}, or \answerNA{}.
    \item[] Justification: Along with the paper, we release a detailed Readme file accompanying the code and trained models. %\justificationTODO{}

\item {\bf Crowdsourcing and research with human subjects}
    \item[] Question: For crowdsourcing experiments and research with human subjects, does the paper include the full text of instructions given to participants and screenshots, if applicable, as well as details about compensation (if any)? 
    \item[] Answer: \answerNA{} % Replace by \answerYes{}, \answerNo{}, or \answerNA{}.
    \item[] Justification: Not applicable. %\justificationTODO{}

\item {\bf Institutional review board (IRB) approvals or equivalent for research with human subjects}
    \item[] Question: Does the paper describe potential risks incurred by study participants, whether such risks were disclosed to the subjects, and whether Institutional Review Board (IRB) approvals (or an equivalent approval/review based on the requirements of your country or institution) were obtained?
    \item[] Answer: \answerNA{} % Replace by \answerYes{}, \answerNo{}, or \answerNA{}.
    \item[] Justification: Not Applicable. %\justificationTODO{}

\item {\bf Declaration of LLM usage}
    \item[] Question: Does the paper describe the usage of LLMs if it is an important, original, or non-standard component of the core methods in this research? Note that if the LLM is used only for writing, editing, or formatting purposes and does \emph{not} impact the core methodology, scientific rigor, or originality of the research, declaration is not required.
    %this research? 
    \item[] Answer: \answerNA{} % Replace by \answerYes{}, \answerNo{}, or \answerNA{}.
    \item[] Justification: We have used an LLM only for paraphrasing text to improve the overall clarity. %\justificationTODO{}

\end{enumerate}

\end{document}